\documentclass{article}
\PassOptionsToPackage{numbers, compress}{natbib}

\usepackage[final]{neurips_2021}

\usepackage[utf8]{inputenc} 
\usepackage[T1]{fontenc}    
\usepackage{hyperref}       
\usepackage{url}            
\usepackage{booktabs}       
\usepackage{amsfonts}       
\usepackage{nicefrac}       
\usepackage{microtype}      
\usepackage{xcolor}         

\usepackage{amsmath}
\usepackage{amssymb}
\usepackage{hyperref}
\usepackage[linesnumbered,ruled,vlined]{algorithm2e}

\SetKwInput{KwInput}{Input}                
\SetKwInput{KwInitialize}{Initialization}                
\SetKwInput{KwOutput}{Output}              

\usepackage{microtype}
\usepackage{graphicx}
\usepackage{subfigure}
\usepackage{booktabs} 
\usepackage{cases}

\usepackage{amsthm}
\newtheorem{definition}{Definition}

\title{Ising Model Selection Using $\ell_{1}$-Regularized Linear Regression: A Statistical Mechanics Analysis}

\author{%
  Xiangming Meng\thanks{Corresponding author.} \\
  Institute for Physics of Intelligence\\
  The University of Tokyo\\
  7-3-1, Hongo, Tokyo 113-0033, Japan \\
  \texttt{meng@g.ecc.u-tokyo.ac.jp} \\
  \And 
  Tomoyuki Obuchi\\
  Department of Systems Science\\
  Kyoto University\\
  Kyoto 606-8501, Japan \\
  \texttt{obuchi@i.kyoto-u.ac.jp} \\
   \AND  
  Yoshiyuki Kabashima\\
  Institute for Physics of Intelligence\\
  The University of Tokyo\\
  7-3-1, Hongo, Tokyo 113-0033, Japan \\
  \texttt{kaba@phys.s.u-tokyo.ac.jp} \\

}

\begin{document}

\maketitle
\begin{abstract}
We theoretically analyze the typical learning performance of $\ell_{1}$-regularized linear regression ($\ell_1$-LinR)
for Ising model selection using the replica method
from statistical mechanics. For typical random regular graphs in the paramagnetic phase, an accurate estimate of the typical sample complexity of $\ell_1$-LinR is obtained.   Remarkably, despite the model misspecification, $\ell_1$-LinR is model selection consistent with the same order of sample complexity as $\ell_{1}$-regularized logistic regression ($\ell_1$-LogR), i.e., $M=\mathcal{O}\left(\log N\right)$,  where $N$ is the number of variables of the Ising model. Moreover, we provide an efficient method to accurately predict the non-asymptotic behavior of $\ell_1$-LinR for moderate $M, N$, such as precision and recall. Simulations show a fairly good agreement between theoretical predictions and experimental results, even for graphs with many loops, which supports our findings. Although this paper mainly focuses on $\ell_1$-LinR, our method is readily applicable for precisely characterizing the typical learning performances of a wide class of  $\ell_{1}$-regularized $M$-estimators including $\ell_1$-LogR and interaction screening. 
\end{abstract}


\section{Introduction}
\vspace{-0.3em}
The advent of massive data across various scientific disciplines has
led to the widespread use of undirected graphical models, also known
as Markov random fields (MRFs), as a tool for discovering and visualizing
dependencies among covariates in multivariate data \citep{wainwright2008graphical}. The Ising model,
originally proposed in statistical physics, is one special class of
binary MRFs with pairwise potentials and has been widely
used in different domains such as image analysis, social networking,
gene network analysis \cite{nguyen2017inverse,aurell2012inverse,bachschmid2015learning,berg2017statistical,bachschmid2017statistical,Abbara2019c}. Among various applications, one fundamental
problem of interest is called Ising model selection, which refers
to recovering the underlying graph structure of the original Ising
model from independent, identically distributed (i.i.d.) samples. 
A variety of methods have been proposed \citep{wainwright2007high,hofling2009estimation,ravikumar2010high,santhanam2012information,decelle2014pseudolikelihood,bresler2015efficiently,vuffray2016interaction, lokhov2018optimal,wu2019sparse, bresler2008reconstruction,prasad2020learning},
demonstrating the possibility of successful Ising model selection
even when the number of samples is smaller than that of the variables.
Notably, it has been demonstrated that for the $\ell_{1}$-regularized logistic regression ($\ell_{1}$-LogR) \citep{ravikumar2010high,wu2019sparse} and interaction screening (IS) \citep{vuffray2016interaction,lokhov2018optimal} estimators,  $M=\mathcal{O}\left(\log N\right)$ samples suffice for an Ising model with $N$ spins under certain assumptions, which is consistent with respect to (w.r.t.) previously established information-theoretic lower-bound \cite{santhanam2012information}. Both $\ell_{1}$-LogR and IS are $\ell_{1}$-regularized $M$-estimators \cite{negahban2012unified} with logistic and IS objective (ISO) loss functions, respectively.

In this paper, we focus on one simpler linear estimator called $\ell_{1}$-regularized linear regression ($\ell_{1}$-LinR) and theoretically investigate its \textit{typical} learning performance using the powerful replica method \citep{mezard1987spin,nishimori2001statistical, opper2001advanced,mezard2009information} from statistical mechanics. The $\ell_{1}$-LinR estimator, widely known as least absolute shrinkage and selection operator (LASSO) \citep{tibshirani1996regression} in statistics and machine learning, is considered here mainly for two reasons. On the one hand, it is one representative example of model misspecification since the quadratic loss of $\ell_{1}$-LinR does not match the true log-conditional-likelihood as $\ell_{1}$-LogR, nor does it 
have the interaction screening property as IS.
On the other hand, as one of the most popular linear estimator, $\ell_{1}$-LinR is more computationally efficient than $\ell_{1}$-LogR and IS, and thus it is of practical importance to investigate its learning performance for Ising model selection. Since it is difficult to obtain results for general graphs, as a first step we consider the random regular (RR) graphs $ \mathcal{G}_{N,d,K_{0}}$ in the paramagnetic phase   \cite{mezard2009information}, where $\mathcal{G}_{N,d,K_{0}}$ denotes the ensemble of RR graphs with constant node degree $d$ and uniform coupling strength $K_0$ on the edges.  


\vspace{-2mm}
\subsection{Contributions}
\vspace{-1mm}
The main contributions are summarized as follows. First, we obtain an accurate estimate of the \textit{typical} sample complexity of  $\ell_1$-LinR for Ising model selection for \textit{typical} RR graphs in the paramagnetic phase, which, remarkably, has the same order as $\ell_1$-LogR. Specifically, for a {typical} RR graph $G\in \mathcal{G}_{N,d,K_{0}}$, using $\ell_1$-LinR with a regularization parameter $0<\lambda<\tanh\left(K_{0}\right)$,  one can consistently reconstruct the structure  with  
$M >\frac{c\left(\lambda,K_{0}\right)\log N}{\lambda^{2}}$ samples, where ${c\left(\lambda,K_{0}\right)=\frac{2\left(1-\tanh^{2}\left(K_{0}\right)+d\lambda^{2}\right)}{1+\left(d-1\right)\tanh^{2}\left(K_{0}\right)}}$. The accuracy of our \textit{typical} sample complexity  prediction is verified by its excellent agreement with experimental results.  To the best of our knowledge, this is the first result that provides an accurate \textit{typical} sample complexity for Ising model selection. 
Interestingly, as $\lambda \to \tanh\left(K_{0}\right)$, a lower bound $M>\frac{2\log N}{\tanh^{2}\left(K_{0}\right)}$ of the typical sample complexity is obtained, which has the same scaling as the information-theoretic  lower bound $M>\frac{c^{\prime}\log N}{K_{0}^{2}}$ \citep{santhanam2012information} for some constant $c^{\prime}$ at high temperatures (i.e., small $K_{0}$) since $\tanh\left(K_{0}\right) = \mathcal{O} \left(K_{0}\right)$ as $K_{0}\rightarrow0$.

Second, we provide a computationally efficient method to precisely predict the typical learning performance of $\ell_{1}$-LinR in the non-asymptotic case with moderate $M, N$, such as  precision, recall, and residual sum of square (RSS). Such precise non-asymptotic predictions of $\ell_{1}$-LinR for Ising model selection have been unavailable even for $\ell_{1}$-LogR \cite{ravikumar2010high,wu2019sparse} and IS \cite{vuffray2016interaction,lokhov2018optimal}, nor are they the same as previous asymptotic results of $\ell_{1}$-LinR assuming fixed $\alpha \equiv M/N$ \citep{bayati2011lasso, rangan2012asymptotic, thrampoulidis2015lasso,gerbelot2020asymptotic}. Moreover, although our theoretical analysis is based on a tree-like structure assumption, experimental results on two dimensional (2D) grid graphs also show a fairly good agreement, indicating that our theoretical result can be a good approximation even for graphs with many loops. 

Third, while this paper mainly focuses on $\ell_{1}$-LinR, our method
is readily applicable to a wide class of  $\ell_{1}$-regularized $M$-estimators \cite{negahban2012unified}, including $\ell_{1}$-LogR \citep{ravikumar2010high} and IS \citep{vuffray2016interaction,lokhov2018optimal}. Thus, an additional technical contribution is providing a generic approach
for precisely characterizing the typical learning performances of various $\ell_{1}$-regularized $M$-estimators for Ising model selection. Although the replica method from statistical mechanics is non-rigorous, our results are conjectured to be correct, which is supported by their excellent agreement with the experimental results.
\textcolor{black}{Additionally, several technical advances we propose in this paper, e.g., the entropy term computation by averaging over the Haar measure and the modification of EOS to address the finite-size effect, might be of general interest to those who use the replica method as a tool for performance analysis.} 


\vspace{-2mm}
\subsection{Related works}
\vspace{-1mm}
There has been some earlier works on the analysis of 
 Ising model selection (also known as the inverse Ising problem) using the replica method \citep{bachschmid2015learning,berg2017statistical,bachschmid2017statistical, Abbara2019c,meng2020structure} from statistical mechanics.
For example, in \citep{bachschmid2017statistical}, the performance of the pseudo-likelihood (PL)
method \citep{besag1975statistical} is studied. However, instead
of graph structure learning, \citep{bachschmid2017statistical} focuses
on the problem of parameter learning. Then, \citep{Abbara2019c} extends the
analysis to the Ising model with sparse couplings using logistic regression
without regularization. The recent work \citep{meng2020structure} analyzes the performance of $\ell_{2}$-regularized linear regression but the techniques invented there are not applicable to  $\ell_{1}$-LinR since the $\ell_{1}$-norm breaks the rotational invariance property.

Regarding the study of $\ell_{1}$-LinR (LASSO) under model misspecification, the past few years have seen a line of research in the field of signal processing with a specific focus on the single-index model \citep{brillinger1982generalized,plan2016generalized,thrampoulidis2015lasso,zhang2016consistency,genzel2016high}. These studies are closely related to ours but there are several important differences. First, in our study, the covariates are generated from an Ising model rather than a Gaussian distribution. Second, we focus on model selection consistency of $\ell_{1}$-LinR while most previous studies consider estimation consistency except \citep{zhang2016consistency}. However,  \citep{zhang2016consistency} only considers the classical asymptotic regime while our analysis  includes the high-dimensional setting where $M\ll N$.

As far as we have searched, there is no earlier study of $\ell_1$-LinR estimator for Ising model selection, though some are found for Gaussian graphical models \citep{meinshausen2006high,zhao2006model}. One closely related work \cite{lokhov2018optimal} states that at high temperatures when the coupling magnitude is approaching zero, both logistic and ISO losses can be approximated by a quadratic loss. However, their claim is only restricted to the very small magnitude near zero while our analysis extends the validity range to the whole paramagnetic phase. Moreover, they evaluate the minimum number of samples necessary for consistently reconstructing ``arbitrary'' Ising models, which, however, seems much larger than that actually needed. By contrast, we provide the first accurate assessment of \textit{typical} sample complexity for consistently reconstructing \textit{typical} samples of Ising models defined over the RR graphs. Furthermore, \cite{lokhov2018optimal} does not provide precise predictions of the non-asymptotic learning performance as we do.

\section{Background and Problem Setup }
\vspace{-0.3em}
\subsection{Ising Model }
\vspace{-1mm}
Ising model is one special class of MRFs with
pairwise potentials and each variable takes binary values \citep{opper2001advanced,mezard2009information},
which is one classical model from statistical physics. The joint probability
distribution of an Ising model with $N$ variables (spins) $\boldsymbol{s}=\left(s_{i}\right)_{i=0}^{N-1}\in\left\{ -1,+1\right\} ^{N}$
has the form
\vspace{-0.8mm}
\begin{equation}
P_{\textrm{Ising}}\left(\boldsymbol{s}|\boldsymbol{J}^*\right)=\frac{1}{Z_{\textrm{Ising}}\left(\boldsymbol{J}^*\right)}\exp\left\{ \sum_{i<j}J^*_{ij}s_{i}s_{j}\right\} ,\label{eq:Ising_distribution}
\end{equation}
where $Z_{\textrm{Ising}}\left(\boldsymbol{J}^*\right)=\sum_{\boldsymbol{s}}\exp\left\{ \sum_{i<j}J^*_{ij}s_{i}s_{j}\right\} $
is the partition function and $\boldsymbol{J}^*=\left(J^*_{ij}\right)_{i,j}$ are the original
couplings, respectively. In general, there are also external fields
but here they are assumed to be zero for simplicity. The structure
of Ising model can be described by an undirected graph $G=\left(\mathtt{V},\mathtt{E}\right) $,
where $\mathtt{V}=\left\{ 0,1,...,N-1\right\}$ is a collection of vertices at which the spins are assigned, and $\mathtt{E}=\left\{ \left(i,j\right)|J^*_{ij}\neq0\right\} $
is a collection of undirected edges, i.e., $J^*_{ij}=0$ for all pairs
of $\left(i,j\right)\notin \mathtt{E}$. For each vertex $i\in \mathtt{V}$, its neighborhood
is defined as the subset $\mathcal{N}\left(i\right)\equiv\left\{ j\in \mathtt{V}|\left(i,j\right)\in \mathtt{E}\right\} $. 
\vspace{-2mm}
\subsection{Neighborhood-based $\ell_{1}$-regularized linear regression ($\ell_{1}$-LinR)}
\vspace{-1mm}
The problem of Ising model selection refers to recovering the graph
$G$ (edge set $\mathtt{E}$), given $M$ i.i.d. samples $\mathcal{D}^{M}=\left\{ \boldsymbol{s}^{\left(1\right)},...,\boldsymbol{s}^{\left(M\right)}\right\}$
from the Ising model. While the maximum likelihood method
has nice properties of consistency and asymptotic efficiency, it suffers
from high computational complexity. To tackle this difficulty, several local learning algorithms have been proposed, notably the $\ell_{1}$-LogR estimator \citep{ravikumar2010high} and IS estimator \citep{vuffray2016interaction}. Both $\ell_{1}$-LogR and IS optimize a regularized local cost function $\ell\left(\cdot\right)$ for each spin
i.e., $\forall i\in \mathtt{V}$, 
\begin{align}
 \boldsymbol{\hat{J}}_{\setminus i} = \underset{\boldsymbol{J}_{\setminus i}}{\arg\min}\left[ \frac{1}{M}\sum_{\mu=1}^{M}\ell\left(s_{i}^{\left(\mu\right)}h_{\setminus i}^{\left(\mu\right)}\right)+\lambda\left\Vert \boldsymbol{J}_{\setminus i}\right\Vert _{1}\right],\label{eq:PL-estimator-df-general}
\end{align}
where $h_{\setminus i}^{\left(\mu\right)}\equiv\sum_{j\neq i}J_{ij}s_{j}^{\left(\mu\right)}$,  $\boldsymbol{J}_{\setminus i} \equiv \left(J_{ij}\right)_{j(\neq i)}$, \textcolor{black}{and $\left\Vert \cdot \right\Vert_1$ denotes the $\ell_1$ norm.}  Specifically,  $\ell\left(x\right)=\log\left(1+e^{-2x}\right)$ for $\ell_{1}$-LogR and $\ell\left(x\right)=e^{-x}$ for IS, which correspond to the minus log conditional distribution \citep{ravikumar2010high} and the ISO \citep{vuffray2016interaction}, respectively. Consequently, the problem of recovering the edge set $\mathtt{E}$ is equivalently reduced to local neighborhood selection, i.e., recovering the neighborhood
set $\mathcal{N}\left(i\right)$
for each vertex $i\in \mathtt{V}$. In particular, given the estimates $\boldsymbol{\hat{J}}_{\setminus i}$ in (\ref{eq:PL-estimator-df-general}),
the neighborhood set of vertex $i$ can be estimated via the nonzero coefficients, i.e., 
\begin{equation}
\hat{\mathcal{N}}\left(i\right)=\left\{ j|\hat{J}_{ij}\neq0,j\in \mathtt{V}\setminus i\right\} ,\;\forall i\in \mathtt{V}.\label{eq:Neighbor_set_estimate}
\end{equation}


In this paper, we focus on one simple linear estimator, termed as the $\ell_{1}$-LinR estimator, i.e., $\forall i\in \mathtt{V}$, 
\begin{align}
 \boldsymbol{\hat{J}}_{\setminus i} = \underset{\boldsymbol{J}_{\setminus i}}{\arg\min}\left[ \frac{1}{2M}\sum_{\mu=1}^{M}\left(s_{i}^{\left(\mu\right)}-h_{\setminus i}^{\left(\mu\right)}\right)^{2}+\lambda\left\Vert \boldsymbol{J}_{\setminus i}\right\Vert _{1}\right],\label{eq:PL-estimator-linear-def}
\end{align}
which, recalling that $s_{i}^{\left(\mu\right)} \in \{-1,+1\}$, corresponds to a quadratic loss $\ell\left(x\right)=\frac{1}{2}\left(x-1\right)^{2}$ in (\ref{eq:PL-estimator-df-general}). The neighorbood set
for each vertex $i\in \mathtt{V}$ is estimated in the same way as (\ref{eq:Neighbor_set_estimate}).
Interestingly, the quadratic loss  used in (\ref{eq:PL-estimator-linear-def})
implies that the postulated conditional distribution is Gaussian and thus inconsistent with the true one, which is one typical case of model misspecification. Furthermore, compared with nonlinear estimators $\ell_1$-LogR and IS, the $\ell_1$-LinR estimator is more efficient to implement. 

\section{\label{Sec3-analysis}Statistical Mechanics Analysis }
\vspace{-0.3em}
\textcolor{black}{In this section, a statistical mechanics analysis of the $\ell_{1}$-LinR estimator is presented for \textit{typical} RR graphs in the paramagnetic phase. Our analysis is applicable to any M-estimator of the form (\ref{eq:PL-estimator-df-general}) and please refer to Appendix \ref{sec:Free-energy-density} for a unified analysis, including detailed results for the $\ell_{1}$-LogR estimator.} 

To characterize the structure learning performance, the {precision} and {recall} are considered:
\begin{equation}
Precision =\frac{TP}{TP+FP}, \;\;\;
Recall =\frac{TP}{TP+FN},\label{eq:Recall-def-1}  
\end{equation}
where $TP$, $FP$, $FN$ denote the number of true positives, false
positives, and false negatives in the estimated couplings, respectively. \textcolor{black}{The concept of \textit{model selection consistent} for an estimator is defined in Definition \ref{def-consistent}, which is also known as the \textit{sparsistency} property \citep{ravikumar2010high}. 
\begin{definition}
An estimator is called model selection consistent if both the associated precision and recall satisfy $Precision\to 1$ and $Recall\to 1$ as $M\rightarrow\infty$. \label{def-consistent}
\end{definition}
Additionally, if one is further interested in the specific values of the estimated couplings, our analysis can also yield the residual sum of squares (RSS) for the estimated couplings.} 

\textcolor{black}{Our theoretical analysis of the learning performance builds on the statistical mechanics framework. Contrary to the probably almost correct (PAC) learning theory \cite{valiant1984theory} in mathematical statistics, statistical mechanics tries to describe the \textit{typical} (as defined in Definition \ref{def-typical}) behavior exactly rather than to bound the \textit{worst case} which is likely to be over-pessimistic \cite{engel2001statistical}. 
\textcolor{black}{
\begin{definition} “typical”  means not just most probable but in addition the probability for situations different from the typical one can be made arbitrarily small as $N \to \infty$ \cite{engel2001statistical}. \label{def-typical}
\end{definition}
}
Similarly, when referring to \textit{typical} RR graphs, we mean tree-like RR graphs, i.e., when seen from a random node, they look like part of an infinite tree, which are  \textit{typical} realizations from the uniform probability distribution on the ensemble of RR graphs.} 

\vspace{-2mm}
\subsection{Problem Formulation}
\vspace{-1mm}
\textcolor{black}{For simplicity and without loss of generality,
we focus on spin $s_{0}$. With a slight abuse of notation, we will drop certain subscripts in following descriptions, e.g.,  $\boldsymbol{J}_{\setminus i}$ will be denoted as $\boldsymbol{J}$ hereafter which represents a vector rather than a matrix.} The basic idea of the statistical mechanical approach is to introduce the following
Hamiltonian and Boltzmann distribution induced by the loss function
$\ell\left(\cdot\right)$
\vspace{-1mm}
\begin{align}
\mathcal{H}\left(\boldsymbol{J}|\mathcal{D}^{M}\right) & =\sum_{\mu=1}^{M}\ell\left(s_{0}^{\left(\mu\right)}h^{\left(\mu\right)}\right)+\lambda M\left\Vert \boldsymbol{J}\right\Vert _{1},\label{eq:Hamilton-def}\\
P\left(\boldsymbol{J}|\mathcal{D}^{M}\right) & =\frac{1}{Z}e^{-\beta\mathcal{H}\left(\boldsymbol{J}|\mathcal{D}^{M}\right)},\label{eq:Boltzmann-def}
\end{align}
where $Z=\int d\boldsymbol{J}e^{-\beta\mathcal{H}\left(\boldsymbol{J}|\mathcal{D}^{M}\right)}$ is
the partition function, and $\beta\left(>0\right)$ is the inverse temperature. 
In the zero-temperature limit $\beta\rightarrow+\infty$, the Boltzmann
distribution (\ref{eq:Boltzmann-def}) converges to a point-wise measure on the estimator (\ref{eq:PL-estimator-df-general}). 
The macroscopic properties of (\ref{eq:Boltzmann-def}) can be analyzed by assessing the free energy density 
$f({\cal D}^M) = - \frac{1}{N\beta} \log Z$, \textcolor{black}{from which, once obtained, we can evaluate averages of various quantities simply by taking its derivatives w.r.t. external fields \cite{nishimori2001statistical}. In  current case,  $f({\cal D}^M)$ depends on the predetermined randomness of ${\cal D}^M$, which plays the role of quenched disorder. As $N, M \to \infty$, $f({\cal D}^M)$ is expected to show the {\em self averaging property} \citep{nishimori2001statistical}}: for typical datasets ${\cal D}^M$, $f({\cal D}^M)$ converges to its average over the random data ${\cal D}^M$:
\vspace{-1mm}
\begin{equation}
f=-\frac{1}{N\beta}\left[\log Z\right]_{\mathcal{D}^{M}},\label{eq:free-energy-def}
\end{equation}
where $\left[\cdot\right]_{\mathcal{D}^{M}}$ denotes expectation over $\mathcal{D}^{M}$, i.e. $\left[\cdot\right]_{\mathcal{D}^{M}}=\sum_{\boldsymbol{s}^{\left(1\right)},...,\boldsymbol{s}^{\left(M\right)}}\left(\cdot\right)\prod_{\mu=1}^{M}P_{\textrm{Ising}}\left(\boldsymbol{s}^{\left(\mu\right)}|\boldsymbol{J}^{*}\right)$.
\textcolor{black}{Consequently, one can analyze the typical performance of any $\ell_1$-regularized M-estimator of the form (\ref{eq:PL-estimator-df-general}) via the assessment of (\ref{eq:free-energy-def}), with $\ell_1$-LinR in (\ref{eq:PL-estimator-linear-def}) being a special case with  $\ell\left(x\right)=\frac{1}{2}\left(x-1\right)^{2}$. }

\vspace{-2mm}
\subsection{\label{subsec:Free-energy-calculation}Replica computation of the
free energy density}
\vspace{-1mm}
Unfortunately, computing (\ref{eq:free-energy-def}) rigorously is difficult. For practically overcoming this difficulty, we resort to the powerful replica method \citep{mezard1987spin,nishimori2001statistical,opper2001advanced, mezard2009information}
from statistical mechanics, which is symbolized using the following
identity 
\begin{equation}
f=-\frac{1}{N\beta}\left[\log Z\right]_{\mathcal{D}^{M}}=-\underset{n\rightarrow0}{\lim}\frac{1}{N\beta}\frac{\partial \log\left[Z^{n}\right]_{\mathcal{D}^{M}}}{\partial n}.\label{eq:replica-def}
\end{equation}
The basic idea is as follows. One replaces the average of $\log Z$ by that of \textcolor{black}{the $n$-th power} $Z^n$ which is analytically tractable for $n \in \mathbb{N}$ in the large $N$ limit, and constructs an analytically continuable expression from $\mathbb{N}$ to $\mathbb{R}$, then takes the limit $n\to 0$ by using the expression. Although the replica method is not rigorous,
it has been empirically verified from extensive studies in disorder
systems \citep{opper2001advanced,mezard2009information} and also found useful in the study of high-dimensional models in machine learning  \citep{gerace2020generalisation,advani2016statistical}. For more details of the replica method, please refer to \citep{mezard1987spin,nishimori2001statistical,opper2001advanced, mezard2009information}. 

Specifically, with the  Hamiltonian $\mathcal{H}\left(\boldsymbol{J}|\mathcal{D}^{M}\right)$, assuming $n\in\mathbb{N}$
is a positive integer, the replicated partition function $\left[Z^{n}\right]_{\mathcal{D}^{M}}$
in (\ref{eq:replica-def}) can be written as 
\begin{align}
\left[Z^{n}\right]_{\mathcal{D}^{M}} & =\int\prod_{a=1}^{n}d\boldsymbol{J}^{a}e^{-\beta\lambda M\sum_{a=1}^{n}\left\Vert \boldsymbol{J}^{a}\right\Vert _{1}}\left\{ \sum_{\boldsymbol{s}}P_{\textrm{Ising}}\left(\boldsymbol{s}|\boldsymbol{J}^{*}\right)\exp\left[-\beta\sum_{a=1}^{n}\ell\left(s_{0}h^{a}\right)\right]\right\} ^{M},\label{eq:Z^n_average_def}
\end{align}
where $h^{a}=\sum_{j}J_{j}^{a}s_{j}$ will be termed as {\em local field} hereafter, \textcolor{black}{and $a$ (and $b$ in the following) is index variable of the replicas}. The analysis below essentially depends on the distribution of $h^{a}$ but it is nontrivial. To resolve it, we  take a similar approach as \citep{Abbara2019c,meng2020structure} and introduce the following ansatz.

\textbf{Ansatz 1 (A1): }\textit{Denote $\Psi=\left\{ j|j\in\mathcal{N}\left(0\right)\right\} $
and $\bar{\Psi}=\left\{ j|j=1,...,N-1,j\notin\mathcal{N}\left(0\right)\right\} $ as the active and inactive sets of spin $s_{0}$, respectively,
then for a typical RR graph $G\in \mathcal{G}_{N,d,K_{0}}$ in the paramagnetic
phase, i.e., }$\left(d-1\right)\tanh^{2}\left(K_{0}\right)<1$\textit{,
the $\ell_{1}$-LinR estimator in (\ref{eq:PL-estimator-linear-def}) \textcolor{black}{ is a random vector determined by random realizations of $\mathcal{D}^{M}$ and }
obeys the following form}
\textit{
\begin{equation}
\hat{J_{j}}=\begin{cases}
\text{\ensuremath{\bar{J}_{j}}}+\frac{1}{\sqrt{N}}w_{j}, & j\in\Psi\\
\frac{1}{\sqrt{N}}w_{j}, & j\in\bar{\Psi}
\end{cases}\label{eq:sparse-anastz}
\end{equation}
where $\text{\ensuremath{\bar{J}_{j}}}$ is the mean value of the
estimator and $w_{j}$ is a random variable which is asymptotically
zero mean with variance scaled as $\mathcal{O}\left(1\right)$.} 

The consistency of Ansatz 1 is checked in Appendix \ref{sec:Verification-of-the}.
Under Ansatz 1, the local fields $h^{a}$ can be decomposed as $h^{a} =\sum_{j\in\Psi}\text{\ensuremath{\bar{J}_{j}}}s_{j}+h_{w}^{a}$
where $h_{w}^{a}\equiv\sum_{j}\frac{1}{\sqrt{N}}w_{j}^{a}s_{j}$ is the ``noise'' part. According
to the central limit theorem (CLT), $h_{w}^{a}$ can be approximated as multivariate Gaussian
variables, which, under the replica symmetric (RS) ansatz \citep{nishimori2001statistical}, can be fully
described by two order parameters
\begin{align}
Q\equiv\frac{1}{N}\sum_{i,j}w_{i}^{a}C_{ij}^{\backslash0}w_{j}^{a},\;\;\;q\equiv\frac{1}{N}\sum_{i,j}w_{i}^{a}C_{ij}^{\backslash0}w_{j}^{b},(a \neq b),\label{eq:Q-q-def}
\end{align}
where $C^{\backslash0}\equiv \{C_{ij}^{\backslash0}\} $
is the covariance matrix of the original Ising model without the spin $s_{0}$. Since the difference between $C^{\backslash 0}$ and that with $s_{0}$ is not essential in the limit $N\to\infty$, hereafter the superscript $^{\backslash0}$ will be discarded.  
As shown in Appendix \ref{sec:Free-energy-density}, \textcolor{black}{for quadratic loss
$\ell\left(x\right)= \frac{1}{2}(1-x)^2$ of $\ell_1$-LinR}, the average free energy density (\ref{eq:replica-def})
in the limit $\beta\rightarrow\infty$ can be computed as
\begin{align}
f\left(\beta\rightarrow\infty\right) & =-\mathtt{Extr}\left\{ -\mathcal{\xi}+S\right\} ,\label{eq:f-result-decomposed}
\end{align}
where $\underset{}{\mathtt{Extr}}\left\{ \cdot\right\}$ denotes the extremum operation w.r.t. relevant variables and $\mathcal{\xi},S$ denote the energy and
entropy terms:
\begin{align}
S & =\textcolor{black}{\underset{\beta\to\infty}{\lim}}\underset{n\rightarrow0}{\lim}\frac{1}{N\beta}\frac{\partial}{\partial n}\log I,\label{eq:S-def}\\
I & =\int\prod_{a=1}^{n}dw^{a} \prod_{a=1}^{n}e^{-\lambda\beta\left\Vert w^{a}\right\Vert _{1}}\delta\left(\sum_{i,j}w_{i}^{a}C_{ij}w_{j}^{a}-NQ\right)
\times \prod_{a<b}\delta\left(\sum_{i,j}w_{i}^{a}C_{ij}w_{j}^{b}-Nq\right),\label{eq:I-def-1}\\
\mathcal{\mathcal{\xi}} & =\frac{\alpha\mathbb{E}_{s,z}\left(s_{0}-\sum_{j\in\Psi}\text{\ensuremath{\bar{J}_{j}}}s_{j}-\sqrt{Q}z\right)^{2}}{2\left(1+\chi\right)}+\alpha\lambda\sum_{j\in\Psi}\left|\bar{J}_{j}\right|,\label{eq:Kesi-result-def}
\end{align}
where $\alpha\equiv M/N, \chi\equiv\lim_{\beta\rightarrow\infty}\beta\left(Q-q\right)$, $\mathbb{E}_{s,z}(\cdot)$ denotes the expectation operation w.r.t. $z \sim {\cal N}(0,1)$ and  $(s_0,\boldsymbol{s}_\Psi) \sim P_{\rm Ising}(s_0,\boldsymbol{s}_\Psi|\boldsymbol{J}^*)\propto e^{s_0\sum_{j\in\Psi}J^{*}_j s_j}$ \citep{Abbara2019c}. \textcolor{black}{For different losses $\ell\left(\cdot\right)$, the free energy results (\ref{eq:f-result-decomposed})  only differ in the energy term $\mathcal{\xi}$, which in general is non-analytical (e.g., logistic loss for $\ell_1$-LogR) but can be solved numerically. Please refer to Appendix \ref{Append-sub-Free energy density result} for more details.}

In contrast to the case of $\ell_{2}$-norm in \citep{meng2020structure},
the $\ell_{1}$-norm in (\ref{eq:I-def-1}) breaks the rotational
invariance property, i.e., $\left\Vert w^{a}\right\Vert _{1}\neq\left\Vert Ow^{a}\right\Vert _{1}$
for general orthogonal matrix $O$, making it difficult to compute
the entropy term $S$. To circumvent this difficulty, we employ an observation that, when considering the RR graph ensemble $\mathcal{G}_{N,d,K_{0}}$ as the coupling network of the Ising model, the orthogonal matrix $O$ diagonalizing the covariance matrix $C$ appears to be distributed from the Haar orthogonal measure \citep{diaconis1994eigenvalues,johansson1997random}. Thus, it is assumed that $I$ in (\ref{eq:I-def-1}) can be replaced by its average $\left[I\right]_{O}$ over the Haar-distributed $O$: 

\textbf{Ansatz 2 (A2): }\textit{Denote $C\equiv\mathbb{E}_{\boldsymbol{s}}[\boldsymbol{s}\boldsymbol{s}^T]$, where $\mathbb{E}_{\boldsymbol{s}}[\cdot] = \sum_{\boldsymbol{s}} P_{\rm Ising}(\boldsymbol{s}|\boldsymbol{J}^*)(\cdot), $
as the covariance matrix of spin configurations $\boldsymbol{s}$.
Suppose that the eigendecomposition of $C$ is $C=O\Lambda O^{T}$,
where $O$ is the orthogonal  matrix, then $O$
can be seen as a random sample generated from the Haar orthogonal
measure and thus for typical graph realizations from $\mathcal{G}_{N,d,K_{0}}$, $I$ in (\ref{eq:I-def-1}) is equal to the average $[I]_O$.}

The consistency of Ansatz (A2) is \textcolor{black}{numerically  checked} in Appendix \ref{sec:Haar-Orthogonal-Assumption}.
Under Ansatz (A2), the entropy term $S$ in (\ref{eq:S-def}) can
be alternatively computed as $\underset{n\rightarrow0}{\lim}\frac{1}{N\beta}\frac{\partial}{\partial n}\log\left[I\right]_{O}$\textit{,
}as shown in Appendix \ref{sec:Free-energy-density}. Finally, under
the RS ansatz, the average free energy density (\ref{eq:replica-def})
in the limit $\beta\rightarrow\infty$ reads
\begin{align}
f\left(\beta\rightarrow\infty\right) = -
 \underset{\mathtt{\varTheta}}{\textrm{Extr}}\left\{ \begin{array}{c}
-\frac{\alpha}{2\left(1+\chi\right)}\mathbb{E}_{s,z}\left(\left(s_{0}-\sum_{j\in\Psi}\text{\ensuremath{\bar{J}_{j}}}s_{j}-\sqrt{Q}z\right)^{2}\right)
-\lambda\alpha\sum_{j\in\Psi}\left|\bar{J}_{j}\right|\\
+\left(-ER+F\eta\right)G^{'}\left(-E\eta\right)
+\frac{1}{2}EQ-\frac{1}{2}F\chi+\frac{1}{2}KR-\frac{1}{2}H\eta \\-\mathbb{E}_{z}\underset{w}{\min}\left\{ \frac{K}{2}w^{2}-\sqrt{H}z w+\frac{\lambda M}{\sqrt{N}}\left|w\right|\right\}
\end{array}\right\} ,\label{eq:free-result-mean}
\end{align}
where $z\sim\mathcal{N}\left(0,1\right)$, and  $G\left(x\right)$ is a function defined as 
\begin{align}
G\left(x\right) & = -\frac{1}{2}\log x-\frac{1}{2} +  \underset{\Lambda}{\mathtt{Extr}}\left\{ -\frac{1}{2}\int\log\left(\Lambda-\gamma\right)\rho\left(\gamma\right)d\gamma+\frac{\Lambda}{2}x\right\},\label{eq:Gx-def-1}
\end{align}
and $\rho\left(\gamma\right)$ is the eigenvalue distribution (EVD)
of the covariance matrix ${C}$, and $\mathtt{\varTheta}$
is a collection of macroscopic parameters $\mathtt{\varTheta}=\left\{ \chi,Q,E,R,F,\eta,K,H,\text{\ensuremath{\left\{  \bar{J}_{j}\right\} } }_{j\in\Psi}\right\}$.
For details of these macroscopic parameters and $\rho\left(\gamma\right)$, please refer
to Appendix \ref{sec:Free-energy-density} and \ref{subsec:Eigenvalue-distribution}, respectively. Note that in (\ref{eq:free-result-mean}),
apart from the ratio $\alpha \equiv M/N$, $N$ and $M$ also appear as $\lambda M/\sqrt{N}$
in the free energy result, which is different from previous results \cite{Abbara2019c,meng2020structure,gerace2020generalisation}.
The reason is that, thanks to the $\ell_{1}$-regularization term
$\lambda M\left\Vert \boldsymbol{J}\right\Vert _{1}$ 
, the mean estimates $\text{\ensuremath{\left\{  \bar{J}_{j}\right\} } }_{j\in\Psi}$
in the active set $\Psi$ and the noise $w$ in the inactive set $\bar{\Psi}$
essentially give different scaling contributions to the free energy
density. 

\textcolor{black}{Although there
are no analytic solutions, these macroscopic parameters in (\ref{eq:free-result-mean}) can be obtained by numerically solving the corresponding equations of state (EOS) employing the physics terminology. Specifically, for the $\ell_1$-LinR estimator, the EOS can be obtained from the extremization condition in  (\ref{eq:free-result-mean}) as follows (for EOS of a general M-esimator and $\ell_1$-LogR, please refer to Appendix \ref{Append-sub-Free energy density result}):}
\begin{equation}
\begin{cases}
E=\frac{\alpha}{\left(1+\chi\right)}, \\
F=\frac{\alpha}{\left(1+\chi\right)^{2}}\left[\mathbb{E}_{s}\left(s_{0}-\sum_{j\in\Psi}s_{j}\bar{J}_{j}\right)^{2}+Q\right], \\
R=\frac{1}{K^{2}}[\left(H+\frac{\lambda^{2}M^{2}}{N}\right)\textrm{erfc}\left(\frac{\lambda M}{\sqrt{2HN}}\right)-2\lambda M\sqrt{\frac{H}{N}}\frac{1}{\sqrt{2\pi}}e^{-\frac{\lambda^{2}M^{2}}{2HN}}], \\
E\eta=-\int\frac{\rho\left(\gamma\right)}{\tilde{\varLambda}-\gamma}d\gamma, \\
Q=\frac{F}{E^{2}}+R\tilde{\varLambda}-\frac{\left(-ER+F\eta\right)\eta}{\int\frac{\rho\left(\gamma\right)}{\left(\tilde{\varLambda}-\gamma\right)^{2}}d\gamma}, \\
K=E\tilde{\varLambda}+\frac{1}{\eta}, \\
\chi=\frac{1}{E}+\eta\tilde{\varLambda}, \\
H=\frac{R}{\eta^{2}}+F\tilde{\varLambda}+\frac{\left(-ER+F\eta\right)E}{\int\frac{\rho\left(\gamma\right)}{\left(\tilde{\varLambda}-\gamma\right)^{2}}d\gamma}, \\
\eta=\frac{1}{K}\textrm{erfc}\left(\frac{\lambda M}{\sqrt{2HN}}\right), \\
\bar{J}_{j}=\frac{\mathtt{soft}\left(\tanh\left(K_{0}\right),\lambda\left(1+\chi\right)\right)}{1+\left(d-1\right)\tanh^{2}\left(K_{0}\right)},j\in\Psi, 
\end{cases}\label{eq:EOS-linear-meanJ-maintext}
\end{equation}
where $\tilde{\varLambda}$ satisfying $E\eta=-\int\frac{\rho\left(\gamma\right)}{\tilde{\varLambda}-\gamma}d\gamma $ is determined by the extremization condition in (\ref{eq:Gx-def-1}) and $\mathtt{soft}\left(z,\tau\right)=\mathtt{sign}\left(z\right)\left(\left|z\right|-\tau\right)_{+}$
is the soft-thresholding function. \textcolor{black}{Once the EOS is solved, the free energy density defined in (\ref{eq:free-energy-def}) is readily obtained.}

\vspace{-2mm}
\subsection{\label{subsec:Asymptotic-statistical-propertie}{High-dimensional asymptotic result}}
\vspace{-1mm}
\vspace{-0.5em}
\begin{figure*}[tbp]
\advance\leftskip -1cm
\centering
\includegraphics[width=13cm]{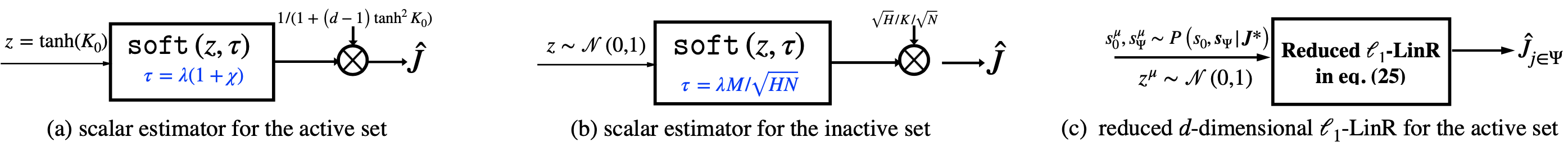}\caption{\small{Equivalent low-dimensional estimators for high-dimensional $\ell_1$-LinR obtained from the statistical mechanics analysis. (a) and (b) are diagrams of the pair of scalar estimators in Eqs. (\ref{eq:J-mean-sub-optimization}) and (\ref{eq:w-optimization}). (c) is a schematic description of the modified estimator in Eq. (\ref{eq:J-mean-sub-optimization-MC}) which takes into account the finite-size effect.    \label{fig:scalar-estimator}}}
\vspace{-1em}
\end{figure*}
One important result of our replica analysis is that, as derived (see Appendix \ref{Append-sub-Free energy density result}) from the free energy result (\ref{eq:free-result-mean}),  the original high dimensional $\ell_{1}$-LinR estimator (\ref{eq:PL-estimator-linear-def}) is \textit{decoupled} into a pair of  scalar estimators, one for the active set and one for the inactive set, i.e., 
\vspace{-0.5mm}
\begin{numcases}{\hat{J}_{j}=}
\frac{\mathtt{soft}\left(\tanh\left(K_{0}\right),\lambda\left(1+\chi\right)\right)}{1+\left(d-1\right)\tanh^{2}\left(K_{0}\right)}\equiv\bar{J}_{j}, & $j\in\Psi$ \label{eq:J-mean-sub-optimization} \\
\frac{\sqrt{H}}{K\sqrt{N}}\mathtt{soft}\left(z_{j},\frac{\lambda M}{\sqrt{HN}}\right), & $j\in\bar{\Psi}$ \label{eq:w-optimization}
\end{numcases}
where $z_j \sim\mathcal{N}\left(0,1\right),j \in \bar{\Psi}$ are i.i.d. standard Gaussian random variables.
The decoupling property  asserts that, once the EOS (\ref{eq:EOS-linear-meanJ-maintext}) is solved, the asymptotic behavior of $\ell_{1}$-LinR can be statistically described by a pair of simple scalar soft-thresholding estimators (see Figs. \ref{fig:scalar-estimator}(a) and \ref{fig:scalar-estimator}(b)).

In the high-dimensional setting where  $N$ is allowed to grow as a function of $M$, one important question is that what is the minimum number of samples $M$ required to achieve model selection consistency as $N\rightarrow\infty$.   Though we obtain a pair of scalar estimators (\ref{eq:J-mean-sub-optimization}) and (\ref{eq:w-optimization}),  there are no analytical solutions to EOS (\ref{eq:EOS-linear-meanJ-maintext}),  making it difficult to derive an explicit condition. To overcome this difficulty, as shown in Appendix \ref{sec:Perfect-graph-recovery}, we perform a perturbation analysis of EOS (\ref{eq:free-result-mean}) and obtain an asymptotic relation $H\simeq F$,
Then, we obtain that for a RR graph $G\in \mathcal{G}_{N,d,K_{0}}$, given $M$ i.i.d. samples $\mathcal{D}^{M}$,
the $\ell_{1}$-LinR estimator (\ref{eq:PL-estimator-linear-def}) can consistently recover the graph structure $G$ as $N\rightarrow\infty$ if 
\begin{align}
M & >\frac{c\left(\lambda,K_{0}\right)\log N}{\lambda^{2}},0<\lambda<\tanh\left(K_{0}\right),\label{eq:C_threshold_lowerbound-1}
\end{align}
where $c(\lambda,K_{0})$ is a constant value dependent on the regularization
parameter $\lambda$ and coupling strength $K_{0}$ and a sharp prediction (as verified in Sec. \ref{sec:experiment}) is obtained as
\begin{equation}
c\left(\lambda,K_{0}\right)=\frac{2\left(1-\tanh^{2}\left(K_{0}\right)+d\lambda^{2}\right)}{1+\left(d-1\right)\tanh^{2}\left(K_{0}\right)}. \label{eq:critical_c_value}
\end{equation}
For details of the analysis, including the counterpart of $\ell_{1}$-LogR, see Appendix \ref{sec:Perfect-graph-recovery}. Consequently, we obtain the \textit{typical} sample complexity of $\ell_1$-LinR for Ising model selection for typical RR graphs in the paramagnetic phase.
The result in (\ref{eq:C_threshold_lowerbound-1}) is derived for $\ell_{1}$-LinR with a fixed regularization parameter $\lambda$. Since the value of $\lambda$ is upper bounded
by $\tanh\left(K_{0}\right)$ (otherwise false negatives occur
as discussed in Appendix \ref{sec:Perfect-graph-recovery}), a lower bound of the typical sample complexity for $\ell_{1}$-LinR is obtained as
\begin{equation}
M>\frac{2\log N}{\tanh^{2}\left(K_{0}\right)}.\label{eq:M-min-result}
\end{equation}
Interestingly, the scaling in  (\ref{eq:M-min-result}) is the same as the information-theoretic worst-case result $M>\frac{c\log N}{K_{0}^{2}}$
obtained in \citep{santhanam2012information} at high temperatures (i.e., small $K_{0}$) since $\tanh\left(K_{0}\right) = \mathcal{O} \left(K_{0}\right)$
as $K_{0}\rightarrow0$. 

\vspace{-2mm}
\subsection{\label{subsec:Statistical-properties-for}{ Non-asymptotic result
for moderate $M,N$}}
\vspace{-1mm}
In practice, it is desirable to predict the non-asymptotic performance of the $\ell_{1}$-LinR estimator for moderate $M,N$. However, the scalar estimator (\ref{eq:J-mean-sub-optimization}) for the active set (see Fig. \ref{fig:scalar-estimator}(a)) fails to capture the fluctuations around the mean estimates.  This is because in obtaining the energy term  $\mathcal{\mathcal{\xi}}$ (\ref{eq:Kesi-result-def}) of the free energy density (\ref{eq:free-result-mean}),
the fluctuations around the mean estimates $\text{\ensuremath{\left\{  \bar{J}_{j}\right\} } }_{j\in\Psi}$
are averaged out by the expectation $\mathbb{E}_{s,z}\left(\cdot\right)$. To address this problem,
we replace $\mathbb{E}_{s,z}\left(\cdot\right)$ in (\ref{eq:free-result-mean}) with a sample average by accounting for the finite-size effect, thus obtaining a modified estimator for the active set as follows
\begin{align}
\text{\ensuremath{\{\hat{J}_{j}\} } }_{j\in\Psi} = \underset{J_{j,j\in\Psi}}{\arg\min}\left[\frac{\sum_{\mu=1}^{M}\left(s_{0}^{\mu}-\sum_{j\in\Psi}s_{j}^{\mu}J_{j}-\sqrt{Q}z^{\mu}\right)^{2}}{2\left(1+\chi\right)M}+\lambda\sum_{j\in\Psi}\left|{J}_{j}\right|\right],\label{eq:J-mean-sub-optimization-MC}
\end{align}
where $s_{0}^{\mu},s_{j,j\in\Psi}^{\mu}\sim P\left(s_{0},\boldsymbol{s}_{\Psi}|\boldsymbol{J}^{*}\right),\;z^{\mu}\sim\mathcal{N}\left(0,1\right),\mu=1...M$. The modified $d$-dimensional estimator   (\ref{eq:J-mean-sub-optimization-MC}) (see Fig. \ref{fig:scalar-estimator}(c) for a schematic) is equivalent to the scalar one (\ref{eq:J-mean-sub-optimization}) (Fig. \ref{fig:scalar-estimator}(a)) as $M\rightarrow\infty$ but it enables
us to capture the fluctuations of $\{  \hat{J}_{j}\}_{j\in\Psi}$
for moderate $M$. Note that due to the replacement of expectation with sample average in  the free energy density (\ref{eq:free-result-mean}), the EOS (\ref{eq:EOS-linear-meanJ-maintext})
also needs to be modified and  it can be solved iteratively as sketched in Algorithm \ref{alg:EOS-finite-L1-alog}.  The details are shown in Appendix \ref{Appendix-Non-asympt-LASSO}.
\begin{algorithm}[H]
 \label{alg:EOS-finite-L1-alog}
\DontPrintSemicolon
  \KwInput{\small{$M,N,\lambda,K_0,\rho\left(\gamma\right)$ and $T_{\rm MC}$}}
  \KwOutput{\small{$\chi,Q,E,R,F,\eta,K,H,\text{\ensuremath{\{\hat{J}^t_{j}\} } }_{j\in\Psi}$}}
  \KwInitialize{\small{$\chi,Q,E,R,F,\eta,K,H$}}
  \small{\textbf{MC sampling}}: \small{For $t=1...T_{MC}$, draw random samples $s_{0}^{\mu,t},\left\{s^{\mu,t}_{j}\right\}_{j\in\Psi}\sim P\left(s_{0},\boldsymbol{s}_{\Psi}|\boldsymbol{J}^{*}\right)$
and $z^{\mu,t}\sim\mathcal{N}\left(0,1\right)$, $\mu=1...M$}

  \Repeat{convergence}{
    \For{$t=1$ {\bfseries to} $T_{\rm MC}$}{
        \small{Solve  $\text{\ensuremath{\{\hat{J}^t_{j}\} } }_{j\in\Psi} = \underset{J_{j,j\in\Psi}}{\arg\min}\left[\frac{\sum_{\mu=1}^{M}\left(s_{0}^{\mu,t}-\sum_{j\in\Psi}s_{j}^{\mu,t}J_{j}-\sqrt{Q}z^{\mu,t}\right)^{2}}{2\left(1+\chi\right)M}+\lambda\sum_{j\in\Psi}\left|{J}_{j}\right|\right]$} 
        
     \small{Compute $\triangle^t=\frac{1}{M}\sum_{\mu=1}^{M}\left(s_{0}^{\mu,t}-\sum_{j\in\Psi}s_{j}^{\mu,t}\hat{J}^t_{j}\right)^{2}$}
    }
    \small{Solve the EOS (\ref{eq:EOS-linear-meanJ-maintext}) with
$\mathbb{E}_{s}\left(s_{i}-\sum_{j\in\Psi}s_{j}\bar{J}_{j}\right)^{2}=\frac{1}{T_{\rm MC}}\sum_{t=1}^{T_{\rm MC}}\triangle^t$}

    \small{Update values of $\chi,Q,E,R,F,\eta,K,H$}
   }

\caption{Method to solve EOS
   (\ref{eq:EOS-linear-meanJ-maintext}) together with (\ref{eq:J-mean-sub-optimization-MC})}
\end{algorithm}

Consequently, for moderate $M,N$, the non-asymptotic statistical properties of the
$\ell_{1}$-LinR estimator can be characterized by the reduced $d$-dimensional $\ell_{1}$-LinR estimator (\ref{eq:J-mean-sub-optimization-MC}) (Fig. \ref{fig:scalar-estimator}(c)) and 
scalar estimator (\ref{eq:w-optimization}) (Fig. \ref{fig:scalar-estimator}(b)) using MC simulations.
Denote $\{ \hat{J}_{j}^{t}\},t=1,...,T_{\rm MC}$ as the estimates in  $t$-th MC simulation, where $\{\hat{J}^{t}_{j}\}_{j\in\Psi}$ and
$\{\hat{J}^{t}_{j}\}_{j\in\bar{\Psi}}$ are solutions of (\ref{eq:J-mean-sub-optimization-MC}) and (\ref{eq:w-optimization}), and 
$T_{\rm MC}$ is the total number of MC simulations. Then,
the $Precision$ and  $Recall$ are computed as
\begin{align}
Precision =\frac{1}{T_{\rm MC}}\sum_{t=1}^{T_{\rm MC}}\frac{\left\Vert \hat{J}_{j,j\in\Psi}^{t}\right\Vert _{0}}{\left\Vert \hat{J}_{j,j\in\Psi}^{t}\right\Vert _{0}+\left\Vert \hat{J}_{j,j\in\bar{\Psi}}^{t}\right\Vert _{0}}, \;\;
Recall =\frac{1}{T_{\rm MC}}\sum_{t=1}^{T_{\rm MC}}\frac{\left\Vert \hat{J}_{j,j\in\Psi}^{t}\right\Vert _{0}}{d},\label{eq:recall}
\end{align}
where $\left\Vert \cdot\right\Vert _{0}$ is the $\ell_0$-norm indicating the number of nonzero elements. In addition, the RSS can be computed as $RSS = \sum_{j}\left|\hat{J}_{j}-{J}^*_{j}\right|^{2} =\frac{1}{T_{\rm MC}}\sum_{t=1}^{T_{\rm MC}}\sum_{j\in\Psi}\left|\hat{J}_{j}^{t}-K_{0}\right|^{2}+R$.

\vspace{-2mm}
\textcolor{black}{
\section{Discussions}
\vspace{-1mm}
It might seem surprising that, despite apparent model misspecification due to the use of quadratic loss, the $\ell_1$-LinR estimator can still correctly infer the structure with the same order of sample complexity as $\ell_1$-LogR. Also, our theoretical analysis implies that the idea of using linear regression for binary data is not as outrageous as one might imagine. Here we provide an intuitive explanation of its success  with a discussion of its limitations.}

\textcolor{black}{
On the average, from (\ref{eq:PL-estimator-linear-def}), the condition for the $\ell_1$-LinR estimator  is given as 
\begin{equation}
\langle s_0s_k\rangle - \sum_{j \ne 0} \langle s_j s_k \rangle J_j  = \lambda \partial |J_k|, \; k = 1,\ldots, N,
\end{equation}
where $\langle\cdot\rangle$ and $\partial |J_k|$ represent average w.r.t. the Boltzmann distribution (\ref{eq:Boltzmann-def}) and the sub-gradient of $|J_k|$, respectively. In the paramagnetic phase, $\langle s_is_j \rangle$ decays in its magnitude exponentially w.r.t.  the distance of sites $i$ and $j$. This guarantees that once the connections $J_k$ of sites in the first nearest neighbor set $\Psi$ are given so that 
\begin{equation}
\langle s_0s_k \rangle= \sum_{j \in \Psi} \langle s_j s_k\rangle J_j + \lambda {\rm{sign}} (J_k), \;\forall{k} \in \Psi \label{conditon-in-Psi}
\end{equation}
holds, the other conditions are automatically satisfied by setting all the other connections that are not from $\Psi$ to zero.  For appropriate choice of $\lambda$, (\ref{conditon-in-Psi}) has solutions of  ${\rm{sign}}(J_k^*)J_k>0, \forall{k} \in \Psi$. Namely $\forall{k} \in \Psi$, the estimate of ${J}_k$ has the same sign as the true value $J_k^*$. This implies that on average the $\ell_1$-LinR estimator can successfully recover the network structure up to the connection signs if $\lambda$ is chosen appropriately.} 

\textcolor{black}{ 
The key of the above argument is that $\langle s_is_j \rangle$ decays exponentially fast w.r.t. the distance of two sites, which does not hold after the phase transition. Thus, it is conjectured that the $\ell_1$-LinR estimator will start to fail in the network recovery just at the phase transition point. However, it is worth noting that this is in fact not  limited to $\ell_1$-LinR: $\ell_1$-LogR also exhibits similar behavior unless post-thresholding is used, as reported in \cite{bento2009graphical}. 
}

\vspace{-2mm}
\section{\label{sec:experiment}Experimental results}
\vspace{-0.3em}
\vspace{-0.8em}
\begin{figure*}[htbp]
\advance\leftskip -1cm
\includegraphics[width=16cm]{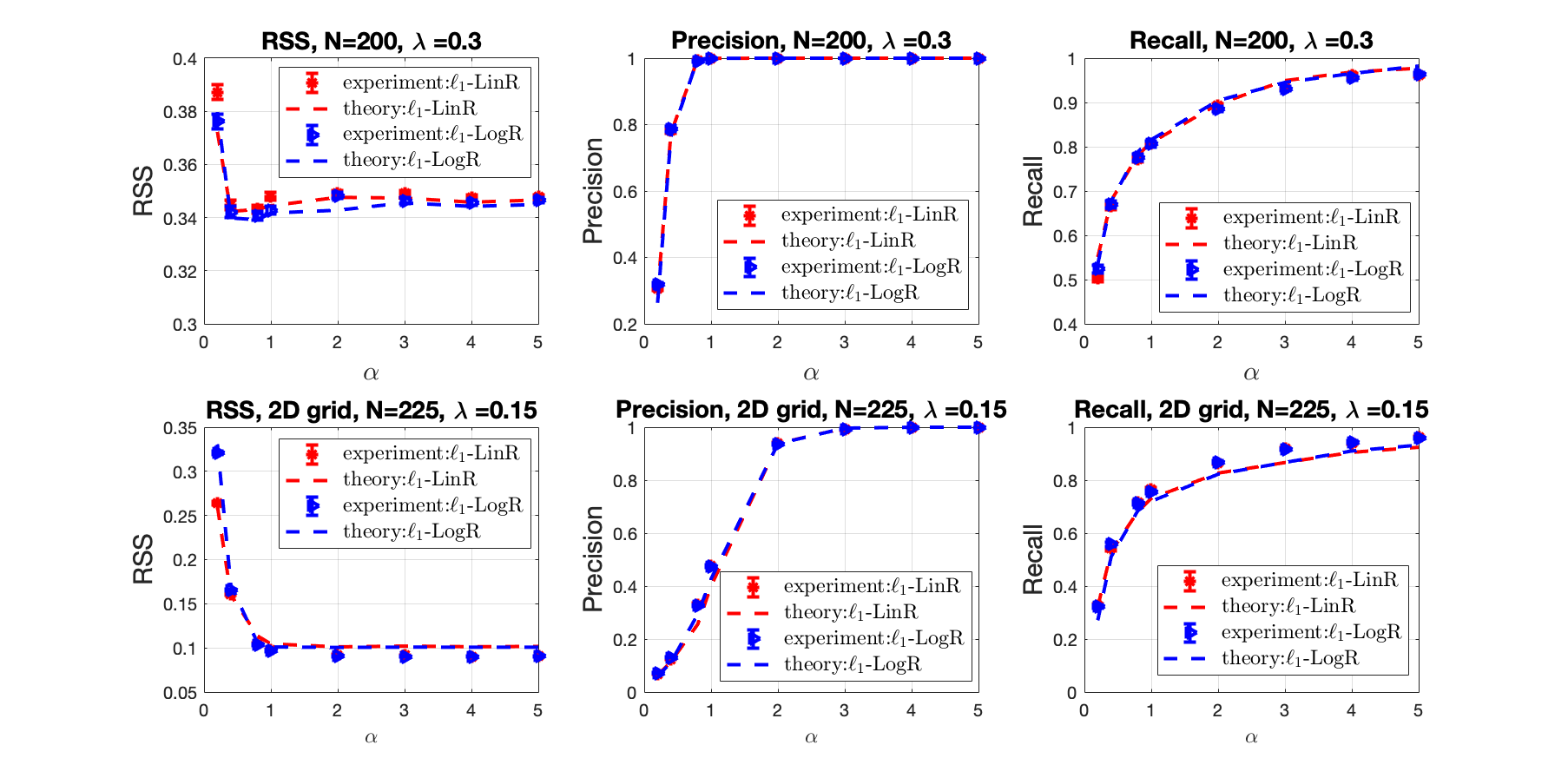}\small{\vspace{-0.8em}\caption{\small{Theoretical and experimental results of RSS, $Precision$ and $Recall$ for RR graph and 2D grid using $\ell_{1}$-LinR and $\ell_{1}$-LogR
with different values of $\alpha \equiv M/N$. The standard error bars are obtained from 1000 random runs. 
A good agreement between theory  and
experiment is achieved, even for small-size 2D grid graph with many loops. For more results, please refer to Appendix \ref{appedix:additional results}.}}\label{fig:RSS-P-R-RR-2D}}
\vspace{-0.5em}
\end{figure*}

\begin{figure*}[htbp]
\begin{center}
\advance\leftskip -1cm
\includegraphics[width=16cm]{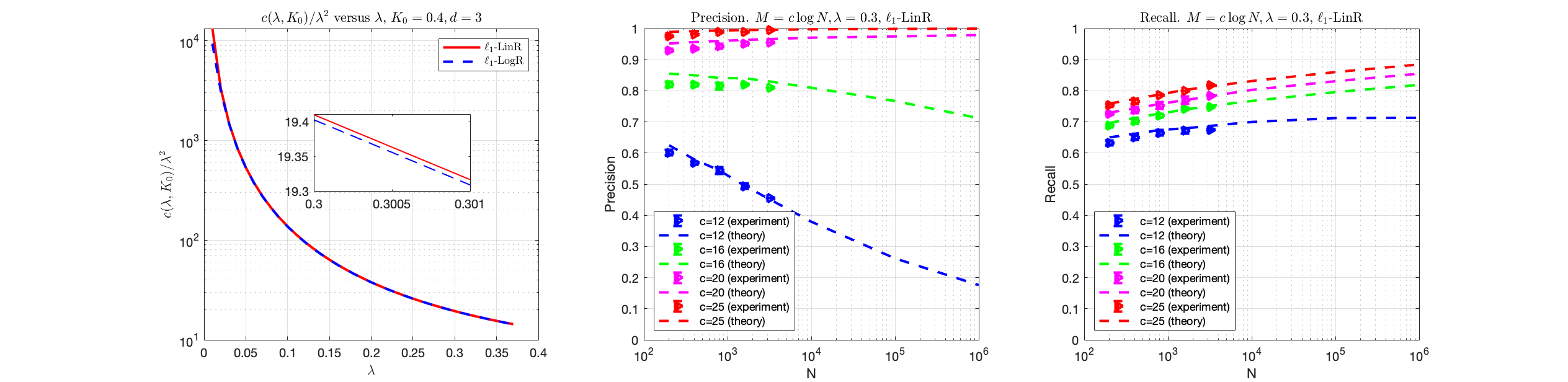}
\small{\caption{\small{\textit{Left}: critical scaling value $c_{0}\left(\lambda,K_{0}\right)\equiv\frac{c\left(\lambda,K_{0}\right)}{\lambda^{2}}$
of $\ell_{1}$-LinR and $\ell_{1}$-LogR for the RR graph $G\in \mathcal{G}_{N,d,K_{0}}$ with $d=3,K_0=0.4$. \textit{Middle and Right}: Precision and Recall for RR graph using $\ell_1$-LinR with $\lambda=0.3$. Experimental results
are shown for $N=200,400,800,1600,3200$. When $c>c_{0}\left(\lambda,K_{0}\right)$ ($c_{0}\left(\lambda=0.3,K_{0}\right)\approx19.41$ in this case),
the \textit{Precision }increases consistently with $N$ and approaches
1 as $N\rightarrow\infty$ while it decreases consistently with $N$
when $c<c_{0}\left(\lambda,K_{0}\right)$. The \textit{Recall} increases
consistently and approach to 1 as $N\rightarrow\infty$. For more results, please refer to Appendix \ref{appedix:additional results}.}}
\label{fig:C_threshold_lambda}}
\vspace{-1.5em}
\end{center}
\end{figure*}

In this section, we conduct numerical experiments to verify the accuracy
of the theoretical analysis. The experimental
procedures are as follows. First, a random graph $G\in \mathcal{G}_{N,d,K_{0}}$
is generated and the Ising model is defined on it. Then, the spin snapshots are obtained \textcolor{black}{
using the Metropolis–Hastings algorithm \citep{metropolis1953equation,hastings1970monte,geman1984stochastic} in the same way as \cite{Abbara2019c}}, yielding the dataset $\mathcal{D}^{M}$.
We randomly choose a center spin $s_{0}$ and infer its neighborhood using the
$\ell_{1}$-LinR (\ref{eq:PL-estimator-linear-def}) and $\ell_{1}$-LogR
\citep{ravikumar2010high} estimators. To obtain standard error bars, we repeat the sequence of operations 1000 times. The RR graph $G\in \mathcal{G}_{N,d,K_{0}}$ with node degree
$d=3$ and coupling strength $K_{0}=0.4$ is considered, which satisfies the paramagnetic
condition $\left(d-1\right)\tanh^{2}\left(K_{0}\right)<1$.
The active couplings $\left\{J_{ij}\right\}_{\left(i,j\right)\in \mathtt{E}}$ have the same
probability of taking both signs of $+1$ or $-1$ \footnote{\footnotesize{Though this setting is different from the analysis where the nonzero couplings take a uniform sign, the result can be directly compared thanks to gauge symmetry \cite{nishimori2001statistical}.}}. 


We first verify the precise non-asymptotic predictions of our method described
in Sec.\ref{subsec:Statistical-properties-for}. Fig. \ref{fig:RSS-P-R-RR-2D} (upper figure)
shows the replica and experimental results of $RSS$, $Precision$, $Recall$ for 
$N=200$ with different values of $\alpha \equiv M/N$. It can be seen that  for both $\ell_{1}$-LinR and $\ell_{1}$-LogR, there is a fairly good agreement between the theoretical predictions and experimental
results, even for small $N=200$ and small $\alpha$ (equivalently
small $M$), verifying the correctness of the replica analysis. Interestingly, a quantitatively similar behavior between $\ell_1$-LinR and $\ell_1$-LogR is observed  \textcolor{black}{in terms of precision and recall. Regarding RSS, the two estimators actually behave differently, which can be clearly seen in Fig. \ref{fig:RSS-Precision-Recall-N200N400N800-lambda01} in Appendix \ref{appedix:additional results}: the RSS is much smaller for $\ell_1$-LogR, which is natural since the  estimates of $\ell_1$-LogR are closer to the true ones due to the model match.} As our theoretical analysis is based on the typical tree-like structure assumption, it is interesting to see if it is applicable to graphs with loops. To this end, we consider the 2D 4-nearest neighbor grid with periodic boundary condition, which is one common loopy graph. Fig. \ref{fig:RSS-P-R-RR-2D} (lower figure)
shows the results for a $15 \times 15$ 2D grid with uniform constant coupling $K_0=0.2$.  The agreement between the theoretical
and numerical results is fairly good, indicating  that our theoretical result can be a
good approximation even for loopy graphs. More results for different values of $N$ and $\lambda$ are shown in  Fig. \ref{fig:RSS-Precision-Recall-N200N400N800-lambda01}
and Fig. \ref{fig:J-Q-R-N200N400} in Appendix \ref{appedix:additional results}.



Subsequently, the asymptotic result and sharpness of the critical scaling value $c_{0}\left(\lambda,K_{0}\right)\equiv\frac{c\left(\lambda,K_{0}\right)}{\lambda^{2}}$ in (\ref{eq:C_threshold_lowerbound-1})
are evaluated. 
First, Fig. \ref{fig:C_threshold_lambda} (left) shows comparison of $c_{0}\left(\lambda,K_{0}\right)$ between $\ell_{1}$-LinR and $\ell_{1}$-LogR for the RR graph
$G\in \mathcal{G}_{N,d,K_{0}}$ when $d=3,K_{0}=0.4$, indicating similar behavior of $\ell_{1}$-LogR and  $\ell_{1}$-LinR. 
Then, we conduct experiments for $M=c\log N$ with different values
of $c$ around $c_{0}\left(\lambda,K_{0}\right)$, and  investigate the trend of $Precision$ and $Recall$ as $N$ increases. When $\lambda=0.3$, Fig. \ref{fig:C_threshold_lambda} (middle and right) show the results of \textit{Precision} and \textit{Recall}, respectively. As expected, the \textit{Precision} increases
consistently with $N$ when
$c>c_{0}\left(\lambda,K_{0}\right)$ and decreases consistently with $N$ when $c<c_{0}\left(\lambda,K_{0}\right)$ while the \textit{Recall} increases consistently and approaches to 1 as $N\rightarrow\infty$, which verifies the sharpness of the critical scaling value prediction.  The results for $\ell_1$-LogR, including the case of $\lambda=0.1$ for both $\ell_1$-LinR and $\ell_1$-LogR,   are shown in  Fig. \ref{fig:Precision-recall-aroundC} and Fig. \ref{fig:Precision-recall-aroundC-1} in Appendix \ref{appedix:additional results}.


\section{Conclusion}
\label{sec:Conclusion}
\vspace{-0.8em}
In this paper, we provide a unified statistical mechanics framework for the analysis of \textit{typical} learning  performances of $\ell_{1}$-regularized $M$-estimators, $\ell_{1}$-LinR in particular, for Ising model selection on typical paramagnetic RR graphs. Using the powerful replica method, the high-dimensional $\ell_{1}$-regularized M-estimator is decoupled into a pair of scalar estimators, by which we obtain an accurate estimate of the typical sample complexity. It is revealed that, perhaps surprisingly, the misspecified $\ell_{1}$-LinR estimator is model selection consistent using $M=\mathcal{O}\left(\log N\right)$ samples, which is of the same order as $\ell_{1}$-LogR. Moreover, with a slight modification of the scalar estimator for the active set to account for the finite-size effect, we further obtain sharp predictions of the non-asymptotic behavior of $\ell_{1}$-LinR (also $\ell_{1}$-LogR) for moderate $M,N$. There is an excellent agreement between theoretical predictions and experimental results, even for graphs with many loops, which supports our findings. Several key assumptions are made in our theoretical analysis, such as the paramagnetic assumption which implies that the coupling strength should not be too large. It is worth noting that the restrictive paramagnetic assumption is not only limited to $\ell_1$-LinR, but also to other low-complexity estimators like $\ell_1$-LogR unless post-thresholding is used \cite{bento2009graphical}.  These assumptions restrict the applicability of the presented
result, and thus overcoming such limitations will be an important
direction for future work.

\section*{Acknowledgements}
This work was supported by JSPS KAKENHI Nos. 17H00764, 18K11463, and 19H01812,
and JST CREST Grant Number JPMJCR1912, Japan.


\bibliography{main}
\bibliographystyle{unsrt}

\newpage
\appendix 
\section{\label{sec:Free-energy-density}Free energy density $f$ computation }
The detailed derivation of the average free energy density $f=-\frac{1}{N\beta}\left[\log Z\right]_{\mathcal{D}^{M}}$
in (\ref{eq:replica-def}) using the replica method is illustrated. Our method provides a unified framework for the statistical mechanics analysis of  any $\ell_1$-regularized $M$-estimator of the form (\ref{eq:PL-estimator-df-general}). As a result, for generality, in the following derivations, we first focus on a generic $\ell_1$-regularized $M$-estimator  (\ref{eq:PL-estimator-df-general}) with a generic loss function $\ell\left(\cdot\right)$. After obtaining the generic results, specific results for both the  $\ell_{1}$-LinR estimator (\ref{eq:PL-estimator-linear-def}) with square
loss $\ell\left(x\right)=\frac{1}{2}\left(x-1\right)^{2}$  and the $\ell_{1}$-LogR estimator with logistic loss $\ell\left(x\right)=\log\left(1+e^{-2x}\right)$ are provided. For the IS estimator, the results can be easily obtained by substituting $\ell\left(x\right)=e^{-x}$, though the specific results are not shown. 

\subsection{Energy term $\xi$ of $f$}
The key of replica method is to compute the replicated partition function
$\left[Z^{n}\right]_{\mathcal{D}^{M}}$. According to the definition in (\ref{eq:Z^n_average_def}) and Ansatz (A1) in Sec. \ref{subsec:Free-energy-calculation},  the average
replicated partition function $\left[Z^{n}\right]_{\mathcal{D}^{M}}$
can be re-written as 
\begin{align}
\left[Z^{n}\right]_{\mathcal{D}^{M}} & =\int\prod_{a=1}^{n}d\boldsymbol{J}^{a}e^{-\beta\lambda M\sum_{a=1}^{n}\sum_{j}\left|J_{j}^{a}\right|}\left\{ \sum_{s}P_{\textrm{Ising}}\left(s|\boldsymbol{J}^{*}\right)\exp\left[-\beta\sum_{a=1}^{n}\ell\left(s_{0}h^{a}\right)\right]\right\} ^{M},\nonumber \\
 & \approx\int\prod_{a=1}^{n}dw^{a}e^{-\beta\lambda M\left(\sum_{a=1}^{n}\sum_{j\in\Psi}\left|\bar{J}_{j}\right|+\sum_{a=1}^{n}\frac{1}{\sqrt{N}}\left\Vert w^{a}\right\Vert _{1}\right)}\times\nonumber \\
 & \left\{ \sum_{s}P_{\textrm{Ising}}\left(s|\boldsymbol{J}^{*}\right)\prod_{a}\int dh_{w}^{a}\delta\left(h_{w}^{a}-\frac{1}{\sqrt{N}}\sum_{j\in\bar{\Psi}}w_{j}^{a}s_{j}\right)e^{-\beta\sum_{a=1}^{n}\ell\left(s_{0}\left(\sum_{j\in\Psi}\text{\ensuremath{\bar{J}_{j}}}s_{j}+h_{w}^{a}\right)\right)}\right\} ^{\alpha N}\nonumber \\
 & =\int\prod_{a=1}^{n}dw^{a}e^{-\beta\lambda M\left(n\sum_{j\in\Psi}\left|\bar{J}_{j}\right|+\sum_{a=1}^{n}\frac{\left\Vert w^{a}\right\Vert _{1}}{\sqrt{N}}\right)}\times\nonumber \\
 & \left\{ \sum_{s_{0},s_{\Psi}} \int \prod_{a=1}^{n}  dh_{w}^{a}  P\left(s_{0},s_{\Psi},\left\{ h_{w}^{a}\right\} _{a}|\boldsymbol{J}^{*},\left\{ w^{a}\right\} _{a}\right)e^{-\beta\sum_{a=1}^{n}\ell\left(s_{0}\left(\sum_{j\in\Psi}\text{\ensuremath{\bar{J}_{j}}}s_{j}+h_{w}^{a}\right)\right)}\right\} ^{\alpha N}\nonumber \\
 & \approx\int\prod_{a=1}^{n}dw^{a}e^{-\beta\lambda M\left(n\sum_{j\in\Psi}\left|\bar{J}_{j}\right|+\sum_{a=1}^{n}\frac{\left\Vert w^{a}\right\Vert _{1}}{\sqrt{N}}\right)}\times\nonumber \\
 & \left\{ \sum_{s_{0},s_{\Psi}}P\left(s_{0},\boldsymbol{s}_{\Psi}|\boldsymbol{J}^{*}\right) \int \prod_{a=1}^{n}  dh_{w}^{a} P_{\textrm{noise}}\left(\left\{ h_{w}^{a}\right\} _{a}|\left\{ w^{a}\right\} _{a}\right)e^{-\beta\sum_{a=1}^{n}\ell\left(s_{0}\left(\sum_{j\in\Psi}\text{\ensuremath{\bar{J}_{j}}}s_{j}+h_{w}^{a}\right)\right)}\right\} ^{\alpha N},\label{eq:Zn-sparse-def}
\end{align}
where  $\left\{ \frac{1}{\sqrt{N}}w_{j}^{a},j\in\Psi \right\}$
in the finite active set $\Psi$ are neglected in the second line when $N$ is large,
$P\left(s_{0},\boldsymbol{s}_{\Psi}|\boldsymbol{J}^{*}\right)=\sum_{\boldsymbol{s}_{\bar{\Psi}}}P_{\textrm{Ising}}\left(s|\boldsymbol{J}^{*}\right)$
is the marginal distribution of $s_{0},\boldsymbol{s}_{\Psi}$ that can be computed
as \citep{Abbara2019c}, $ P_{\textrm{noise}}\left(\left\{ h_{w}^{a}\right\} _{a}|\left\{ w^{a}\right\} _{a}\right)$
is the  distribution  of the ``noise'' part $h_{w}^{a} \equiv\frac{1}{\sqrt{N}}\sum_{j\in\bar{\Psi}}w_{j}^{a}s_{j}$ of the local field. In the last line, the asymptotic independence between $h_{w}^{a}$ and ($s_{0},\boldsymbol{s}_{\Psi}$) are applied as discussed in \citep{Abbara2019c}.   \textcolor{black}{Regarding the marginal distribution $P\left(s_{0},\boldsymbol{s}_{\Psi}|\boldsymbol{J}^{*}\right)$, in general we have to take into account the cavity fields in the marginal distribution. In the case considered in this paper, however, the paramagnetic assumption simplifies the marginal distribution and finally it is proportional to $e^{s_0\sum_{j\in\Psi}J_{j}^{*} s_j}$  \citep{Abbara2019c}. When $\Psi$ has a small cardinality $d$, we can compute the expectation w.r.t. $(s_0, s_{\Psi})$ exactly by exhaustive enumeration. For large $d$, MC methods like the Metropolis–Hastings algorithm \citep{metropolis1953equation,hastings1970monte,geman1984stochastic} might be used.}

To proceed with the calculation, according to the CLT,
the noise part $\left\{ h_{w}^{a}\right\} _{a=1}^{n}$ can
be regarded as Gaussian variables so that
$ P_{\textrm{noise}}\left(\left\{ h_{w}^{a}\right\} _{a}|\left\{ w^{a}\right\} _{a}\right)$
can be approximated as a multivariate Gaussian distribution. Under
the RS ansatz, two auxiliary order parameters
are introduced, i.e.,
\begin{align}
Q \equiv& \frac{1}{N}\sum_{i,j\in\bar{\Psi}}w_{i}^{a}C_{ij}^{\backslash0}w_{j}^{a},\label{eq:Q-def}\\
q \equiv & \frac{1}{N}\sum_{i,j\in\bar{\Psi}}w_{i}^{a}C_{ij}^{\backslash0}w_{j}^{b},\; \left(a \neq b\right),\label{eq:q-def}
\end{align}
where $C^{\backslash0}=\left\{ C_{ij}^{\backslash0}\right\} $
is the covariance matrix of the original Ising model without $s_{0}$. To write the integration in
terms of the order parameters $Q,q$, we introduce the following trivial
identities
\begin{align}
1 & =N\int dQ\delta\left(\sum_{i,j\neq0}w_{i}^{a}C_{ij}^{\backslash0}w_{j}^{a}-NQ\right),a=1,...,n\\
1 & =N\int dq\delta\left(\sum_{i,j\neq0}w_{i}^{a}C_{ij}^{\backslash0}w_{j}^{b}-Nq\right),a<b,\label{eq:trivial-delta-func}
\end{align}
so that $\left[Z^{n}\right]_{\mathcal{D}^{M}}$ in (\ref{eq:Zn-sparse-def})
can be rewritten as 
\begin{align}
\left[Z^{n}\right]_{\mathcal{D}^{M}} & =e^{-\beta\lambda Mn\sum_{j\in\Psi}\left|\bar{J}_{j}\right|}\int dQdq\int\prod_{a=1}^{n}dw^{a}e^{-\lambda\beta\frac{M}{\sqrt{N}}\sum_{a=1}^{n}\left\Vert w^{a}\right\Vert _{1}} \nonumber \\
&\times \prod_{a=1}^{n}\delta\left(\sum_{i,j}w_{i}^{a}C_{ij}^{\backslash0}w_{j}^{a}-NQ\right)
 \prod_{a<b}\delta\left(\sum_{i,j}w_{i}^{a}C_{ij}^{\backslash0}w_{j}^{b}-Nq\right)\times\nonumber \\
 & \left\{ \sum_{s_{0},s_{\Psi}}P\left(s_{0},\boldsymbol{s}_{\Psi}|\boldsymbol{J}^{*}\right)\int\prod_{a=1}^{n}dh_{w}^{a} P_{\textrm{noise}}\left(\left\{ h_{w}^{a}\right\} _{a}|\left\{ w^{a}\right\} _{a}\right)e^{-\beta\sum_{a=1}^{n}\ell\left(s_{0}\left(\sum_{j\in\Psi}\text{\ensuremath{\bar{J}_{j}}}s_{j}+h_{w}^{a}\right)\right)}\right\} ^{\alpha N}\\
= & \int dQdqIe^{M\log L},
\end{align}
where
\begin{align}
I & \equiv\int\prod_{a=1}^{n}dw^{a}e^{-\lambda\beta\frac{M}{\sqrt{N}}\sum_{a=1}^{n}\left\Vert w^{a}\right\Vert _{1}}\prod_{a=1}^{n}\delta\left(\sum_{i,j}w_{i}^{a}C_{ij}^{\backslash0}w_{j}^{a}-NQ\right)\prod_{a<b}\delta\left(\sum_{i,j}w_{i}^{a}C_{ij}^{\backslash0}w_{j}^{b}-Nq\right),\label{eq:NS-part-sparse}\\
L & \equiv e^{-\beta\lambda n\sum_{j\in\Psi}\left|\bar{J}_{j}\right|}\sum_{s_{0},\boldsymbol{s}_{\Psi}}P\left(s_{0},\boldsymbol{s}_{\Psi}|\boldsymbol{J}^{*}\right)\int\prod_{a=1}^{n}dh_{w}^{a} P_{\textrm{noise}}\left(\left\{ h_{w}^{a}\right\} _{a}|\left\{ w^{a}\right\} _{a}\right)e^{-\beta\sum_{a=1}^{n}\ell\left(s_{0}\left(\sum_{j\in\Psi}\text{\ensuremath{\bar{J}_{j}}}s_{j}+h_{w}^{a}\right)\right)}.\label{eq:L-part-df-1}
\end{align}

According to CLT and (\ref{eq:Q-def}) and (\ref{eq:q-def}), the noise parts $h_{w}^{a},a=1,\ldots,n$ follow a
multivariate Gaussian distribution with zero mean (paramagnetic assumption)
and covariances 
\begin{equation}
\left\langle h_w^a h_w^b\right\rangle ^{\backslash0}=Q\delta_{ab}+\left(1-\delta_{ab}\right)q. \label{eq:cov-h-def}
\end{equation}

Consequently, by introducing  two auxiliary i.i.d. standard Gaussian random
variables $v_{a}\sim\mathcal{N}\left(0,1\right),z \sim\mathcal{N}\left(0,1\right) $, the
noise parts $h_{w}^{a},a=1,\ldots,n$ can be written
in a compact form
\begin{equation}
h_{w}^a=\sqrt{Q-q}v_{a}+\sqrt{q}z,a=1,\ldots,n\label{eq:local-field-compactform}
\end{equation}
so that $L$ in (\ref{eq:L-part-df-1}) could be written
as
\begin{align}
L & =e^{-\beta\lambda n\sum_{j\in\Psi}\left|\bar{J}_{j}\right|}\sum_{s_{0},\boldsymbol{s}_{\Psi}}P\left(s_{0},\boldsymbol{s}_{\Psi}|\boldsymbol{J}^{*}\right)\int\prod_{a=1}^{n}dh_{w}^{a} P_{\textrm{noise}}\left(\left\{ h_{w}^{a}\right\} _{a}|\left\{ w^{a}\right\} _{a}\right)e^{-\beta\sum_{a=1}^{n}\ell\left(s_{0}\left(\sum_{j\in\Psi}\text{\ensuremath{\bar{J}_{j}}}s_{j}+h_{w}^{a}\right)\right)}\nonumber \\
 & =e^{-\beta\lambda n\sum_{j\in\Psi}\left|\bar{J}_{j}\right|}\sum_{s_{0},\boldsymbol{s}_{\Psi}}P\left(s_{0},\boldsymbol{s}_{\Psi}|\boldsymbol{J}^{*}\right)\int\mathcal{D}z\prod_{a}\mathcal{D}v_{a}e^{-\beta\sum_{a=1}^{n}\ell\left(s_{0}\left(\sum_{j\in\Psi}\text{\ensuremath{\bar{J}_{j}}}s_{j}+\sqrt{Q-q}v_{a}+\sqrt{q}z\right)\right)}\nonumber \\
 & =e^{-\beta\lambda n\sum_{j\in\Psi}\left|\bar{J}_{j}\right|}\sum_{s_{0},\boldsymbol{s}_{\Psi}}P\left(s_{0},\boldsymbol{s}_{\Psi}|\boldsymbol{J}^{*}\right)\int\mathcal{D}z\left[\underset{A}{\underbrace{\int\mathcal{D}ve^{-\beta\ell\left(s_{0}\left(\sum_{j\in\Psi}\text{\ensuremath{\bar{J}_{j}}}s_{j}+\sqrt{Q-q}v+\sqrt{q}z\right)\right)}}}\right]^{n}\nonumber \\
 & =e^{-\beta\lambda n\sum_{j\in\Psi}\left|\bar{J}_{j}\right|}\sum_{s_{0},\boldsymbol{s}_{\Psi}}P\left(s_{0},\boldsymbol{s}_{\Psi}|\boldsymbol{J}^{*}\right)\mathbb{E}_{z}\left(A^{n}\right),
\end{align}
where $\mathcal{D}z=\frac{dz}{\sqrt{2\pi}}e^{-\frac{z^{2}}{2}}$.
As a result, using the replica formula, we have
\begin{align}
&\underset{n\rightarrow0}{\lim}\frac{1}{n}\log L \nonumber \\ =&-\beta\lambda\sum_{j\in\Psi}\left|\bar{J}_{j}\right|+\underset{n\rightarrow0}{\lim}\frac{\log\sum_{s_{0},\boldsymbol{s}_{\Psi}}P\left(s_{0},\boldsymbol{s}_{\Psi}|\boldsymbol{J}^{*}\right)E_{z}\left(A^{n}\right)}{n}\nonumber \\
 =&-\beta\lambda\sum_{j\in\Psi}\left|\bar{J}_{j}\right|+\mathbb{E}_{z}\left[\sum_{s_{0},\boldsymbol{s}_{\Psi}}P\left(s_{0},\boldsymbol{s}_{\Psi}|\boldsymbol{J}^{*}\right)\log A\right]\nonumber \\
 =&-\beta\lambda\sum_{j\in\Psi}\left|\bar{J}_{j}\right|+\sum_{s_{0},\boldsymbol{s}_{\Psi}}P\left(s_{0},\boldsymbol{s}_{\Psi}|\boldsymbol{J}^{*}\right)\int\mathcal{D}z\log\int\mathcal{D}ve^{-\beta\ell\left(s_{0}\left(\sum_{j\in\Psi}\text{\ensuremath{\bar{J}_{j}}}s_{j}+\sqrt{Q-q}v+\sqrt{q}z\right)\right)}\nonumber \\
 =&-\beta\lambda\sum_{j\in\Psi}\left|\bar{J}_{j}\right|+\sum_{s_{0},\boldsymbol{s}_{\Psi}}P\left(s_{0},\boldsymbol{s}_{\Psi}|\boldsymbol{J}^{*}\right)\int\mathcal{D}z\log\int\frac{dy}{\sqrt{2\pi\left(Q-q\right)}}e^{-\frac{\left[y-s_{0}\left(\sum_{j\in\Psi}\text{\ensuremath{\bar{J}_{j}}}s_{j}+\sqrt{q}z\right)\right]^{2}}{2\left(Q-q\right)}}e^{-\beta\ell\left(y\right)},\label{eq:logL/n-result}
\end{align}
where in the last line, a change of variable $y=s_{0}\left(\sum_{j\in\Psi}\text{\ensuremath{\bar{J}_{j}}}s_{j}+\sqrt{Q-q}v+\sqrt{q}z\right)$
is used. 

As a result, from (\ref{eq:replica-def}), the average
free energy density in the limit $\beta\rightarrow\infty$ reads 
\begin{align}
f\left(\beta\rightarrow\infty\right) & =\lim_{\beta\rightarrow\infty}-\frac{1}{N\beta}\left\{ \underset{n\rightarrow0}{\lim}\frac{\partial}{\partial n}\log I+M\underset{n\rightarrow0}{\lim}\frac{\partial}{\partial n}\log L\right\} \nonumber \\
 & =-\mathtt{Extr}\left\{ -\mathcal{\xi}+S\right\} ,\label{eq:free-energy-result-1}
\end{align}
where ${\textrm{Extr}}\left\{ \cdot\right\} $ denotes
extremization w.r.t. some relevant variables, and $\mathcal{\xi},S$ are the corresponding energy and
entropy terms of $f$, respectively:  
\begin{align}
S & =\underset{n\rightarrow0}{\lim}\frac{1}{N\beta}\frac{\partial}{\partial n}\log I,\label{eq:S-def-1}\\
I & =\int\prod_{a=1}^{n}dw^{a}e^{-\lambda\beta\sum_{a=1}^{n}\left\Vert w^{a}\right\Vert _{1}}\prod_{a=1}^{n}\delta\left(\sum_{i,j}w_{i}^{a}C_{ij}w_{j}^{a}-NQ\right)\prod_{a<b}\delta\left(\sum_{i,j}w_{i}^{a}C_{ij}w_{j}^{b}-Nq\right),\label{eq:I-def}\\
\mathcal{\mathcal{\xi}} & =\alpha\lambda\sum_{j\in\Psi}\left|\bar{J}_{j}\right|+\alpha \mathbb{E}_{s,z}\left( \underset{y}{\min}\left[\frac{\left(y-s_{0}\left(\sqrt{Q}z+\sum_{j\in\Psi}\text{\ensuremath{\bar{J}_{j}}}s_{j}\right)\right)^{2}}{2\chi}+\ell\left(y\right)\right]\right),\label{eq:Kesi-result}
\end{align}
and the relation $\lim_{\beta\rightarrow\infty}\beta\left(Q-q\right)\equiv\chi$
is used \citep{bachschmid2017statistical,Abbara2019c}. \textcolor{black}{The extremization in the free energy result (\ref{eq:free-energy-result-1}) comes from saddle point method in the large $N$ limit.}  

\subsection{Entropy term $S$ of $f$ }
To obtain the final result of free energy density, there is still one remaining entropy 
term $S$ to compute, which requires the result of $I$ (\ref{eq:I-def}).
However, unlike the $\ell_{2}$-norm, the $\ell_{1}$-norm  in (\ref{eq:I-def}) breaks the rotational invariance property, which makes the computation of $I$ difficult and the methods in \citep{Abbara2019c,meng2020structure} are no loner applicable.  To address this problem, applying the Haar Orthogonal
Ansatz (A2) in Sec. \ref{subsec:Free-energy-calculation}, we employ  a method to replace $I$ with an average $\left[I\right]_{O}$ over the orthogonal  matrix
$O$ generated from the Haar orthogonal measure. 

Specifically, also under the RS ansatz,  two auxiliary order parameters
are introduced, i.e.,
\begin{align}
R \equiv& \frac{1}{N}\sum_{i,j}w_{i}^{a}w_{j}^{a},\label{eq:R-def}\\
r \equiv & \frac{1}{N}\sum_{i,j}w_{i}^{a}w_{j}^{b},\; \left(a \neq b\right).\label{eq:r-def}
\end{align}
Then, by inserting the delta functions $\prod_{a}\delta\left(\left(w^{a}\right)^{T}w^{a}-NR\right)\prod_{a<b}\delta\left(\left(w^{a}\right)^{T}w^{b}-Nr\right)$, we obtain 
\begin{align}
I & =\int\prod_{a=1}^{n}dw^{a}e^{-\frac{\lambda\beta M}{\sqrt{N}}\sum_{a=1}^{n}\left\Vert w^{a}\right\Vert _{1}}\prod_{a=1}^{n}\delta\left(\left(w^{a}\right)^{T}Cw^{a}-NQ\right)\prod_{a<b}\delta\left(\left(w^{a}\right)^{T}Cw^{b}-Nq\right)\nonumber \\
 & \times\int dRdr\prod_{a}\delta\left(\left(w^{a}\right)^{T}w^{a}-NR\right)\prod_{a<b}\delta\left(\left(w^{a}\right)^{T}w^{b}-Nr\right).\label{eq:V_form2}
\end{align}

Moreover, replacing the original delta functions in (\ref{eq:V_form2})
as the following identities
\[
\begin{cases}
\delta\left(\left(w^{a}\right)^{T}Cw^{a}-NQ\right)=\int d\hat{Q}e^{-\frac{\hat{Q}}{2}\left(\left(w^{a}\right)^{T}Cw^{a}-NQ\right)},\\
\delta\left(\left(w^{a}\right)^{T}Cw^{b}-Nq\right)=\int d\hat{q}e^{\hat{q}\left(\left(w^{a}\right)^{T}Cw^{b}-Nq\right)},
\end{cases}
\]
and taking average over the orthogonal matrix $O$, after some algebra, the $I$ is replaced with the following average $\left[I\right]_{O}$
\begin{align}
\left[I\right]_{O} & =\int dRdrd\hat{Q}d\hat{q}\prod_{a=1}^{n}dw^{a}e^{-\frac{\lambda\beta M}{\sqrt{N}}\sum_{a=1}^{n}\left\Vert w^{a}\right\Vert _{1}}\prod_{a}\delta\left(\left(w^{a}\right)^{T}w^{a}-NR\right)\prod_{a<b}\delta\left(\left(w^{a}\right)^{T}w^{b}-Nr\right)\nonumber \\
 & \times\exp\left\{ \frac{Nn}{2}\hat{Q}Q-\frac{Nn}{2}\left(n-1\right)\hat{q}q\right\} \times\left[e^{\frac{1}{2}\mathtt{Tr}\left(CL_{n}\right)}\right]_{O},\\
L_{n} & =-\left(\hat{Q}+\hat{q}\right)\sum_{a=1}^{n}w^{a}\left(w^{a}\right)^{T}+\hat{q}\left(\sum_{a=1}^{n}w^{a}\right)\left(\sum_{b=1}^{n}w^{b}\right)^{T}.
\end{align}
To proceed with the computation, the eigendecompostion of the matrix  $L_{n}$ is performed. After some algebra, for the configuration of $w^{a}$ that satisfies both $\left(w^{a}\right)^{T}w^{a}=NR$
and $\left(w^{a}\right)^{T}w^{b}=Nr$, the eigenvalues
and associated eigenvectors of matrix $L_{n}$ can be calculated as
follows
\begin{align}
\begin{cases}
\lambda_{1}=-N\left(\hat{Q}+\hat{q}-n\hat{q}\right)\left(R-r+nr\right),\\
u_{1}=\sum_{a=1}^{n}w^{a},\\
\lambda_{2}=-N\left(\hat{Q}+\hat{q}\right)\left(R-r\right),\\
u_{a}=w^{a}-\frac{1}{n}\sum_{b=1}^{n}w^{b},a=2,...,n,
\end{cases}\label{eq:eigenvalue-eigenvector-Ln}
\end{align}
where $\lambda_{1}$ is the eigenvalue corresponding to the eigenvector $u_1$ while $\lambda_{2}$ is the degenerate  eigenvalue corresponding to eigenvectors $u_a,a=2,...,n$. 
To compute $\left[e^{\frac{1}{2}\mathtt{Tr}\left(CL_{n}\right)}\right]_{O}$, we define a function $G\left(x\right)$
as 
\begin{align}
G\left(x\right) & \equiv\frac{1}{N}\log\left[\exp\left(\frac{x}{2}\mathtt{Tr}C\left(\mathbf{1}\mathbf{1}^{T}\right)\right)\right]_{O}\nonumber \\
 & =\underset{\Lambda}{\mathtt{Extr}}\left\{ -\frac{1}{2}\int\log\left(\Lambda-\gamma\right)\rho\left(\gamma\right)d\gamma+\frac{\Lambda}{2}x\right\} -\frac{1}{2}\log x-\frac{1}{2},\label{eq:Gx-def}
\end{align}
and $\rho\left(\gamma\right)$ is the eigenvalue distribution (EVD)
of $C$. Then, combined with (\ref{eq:eigenvalue-eigenvector-Ln}),
after some algebra, we obtain that 
\begin{equation}
\frac{1}{N}\log\left[e^{\frac{1}{2}\mathtt{Tr}\left(CL_{n}\right)}\right]_{O}=G\left(-\left(\hat{Q}+\hat{q}-n\hat{q}\right)\left(R-r+nr\right)\right)+\left(n-1\right)G\left(-\left(\hat{Q}+\hat{q}\right)\left(R-r\right)\right).
\end{equation}

Furthermore, replacing the original delta functions in (\ref{eq:V_form2})
as 
\[
\begin{cases}
\delta\left(\left(w^{a}\right)^{T}w^{a}-NR\right)=\int d\hat{R}e^{-\frac{\hat{R}}{2}\left(\left(w^{a}\right)^{T}w^{a}-NR\right)},\\
\delta\left(\left(w^{a}\right)^{T}w^{b}-Nr\right)=\int d\hat{r}e^{\hat{r}\left(\left(w^{a}\right)^{T}w^{b}-Nr\right)},
\end{cases}
\]
we obtain 
\begin{align}
\left[I\right]_{0} & =\int dRdrd\hat{Q}d\hat{q}d\hat{R}d\hat{r}\prod_{a=1}^{n}dw^{a}\exp\left\{ -\sum_{a=1}^{n}\frac{\lambda\beta M}{\sqrt{N}}\left\Vert w^{a}\right\Vert _{1}-\frac{\hat{R}+\hat{r}}{2}\sum_{a=1}^{n}\left(w^{a}\right)^{T}w^{a}+\frac{\hat{r}}{2}\sum_{a,b}\left(w^{a}\right)^{T}w^{b}\right\} \nonumber \\
 & \times\exp\left\{ \frac{Nn}{2}\hat{R}R-\frac{Nn}{2}\left(n-1\right)\hat{r}r+\frac{Nn}{2}\hat{Q}Q-\frac{Nn}{2}\left(n-1\right)\hat{q}q\right\} \times\left[e^{\frac{1}{2}\mathtt{Tr}\left(CL_{n}\right)}\right]_{O}.
\end{align}
In addition, using a Gaussian integral, the following result can be linearized as 
\begin{align*}
 & \int\prod_{a=1}^{n}dw^{a}\exp\left\{ -\sum_{a=1}^{n}\frac{\lambda\beta M}{\sqrt{N}}\left\Vert w^{a}\right\Vert _{1}-\frac{\hat{R}+\hat{r}}{2}\sum_{a=1}^{n}\left(w^{a}\right)^{T}w^{a}+\frac{\hat{r}}{2}\sum_{a,b}\left(w^{a}\right)^{T}w^{b}\right\} \\
= & \int \prod_{a=1}^{n}dw^{a}\exp\left\{ -\sum_{a=1}^{n}\sum_{i=1}^{N}\frac{\lambda\beta M}{\sqrt{N}}\left|w_{i}^{a}\right|-\frac{\hat{R}+\hat{r}}{2}\sum_{a=1}^{n}\sum_{i=1}^{N}\left(w_{i}^{a}\right)^{2}+\frac{\hat{r}}{2}\sum_{i=1}^{N}\left(\sum_{a=1}^{n}w_{i}^{a}\right)^{2}\right\} \\
= &\prod_{i}\int\mathcal{D}z_{i}\int\prod_{a=1}^{n}dw^{a}\exp\left\{ -\sum_{a=1}^{n}\frac{\lambda\beta M}{\sqrt{N}}\left|w_{i}^{a}\right|-\frac{\hat{R}+\hat{r}}{2}\sum_{a=1}^{n}\left(w_{i}^{a}\right)^{2}+\sqrt{\hat{r}}z_{i}\sum_{a}w_{i}^{a}\right\} \\
= & \prod_{i}\int\mathcal{D}z_{i}\left\{ \int dw\exp\left[-\frac{\hat{R}+\hat{r}}{2}w_{i}^{2}+\left(\sqrt{\hat{r}}z-\frac{\lambda\beta M}{\sqrt{N}}\textrm{ sign}\left(w_{i}\right)\right)w_{i}\right]\right\} ^{n},
\end{align*}
where $\mathcal{D}{z_i}=\frac{dz_i}{\sqrt{2\pi}}e^{-\frac{{z_i}^{2}}{2}}$. Consequently, the entropy term $S$ of the free energy density $f$ is computed as
\begin{align*}
\underset{n\rightarrow0}{\lim}\frac{1}{N}\frac{\partial}{\partial n}\log\left[I\right]_{O} & =\left(\hat{q}\left(R-r\right)-\left(\hat{Q}+\hat{q}\right)r\right)G^{'}\left(-\left(\hat{Q}+\hat{q}\right)\left(R-r\right)\right) \\
& +G\left(-\left(\hat{Q}+\hat{q}\right)\left(R-r\right)\right)+\frac{\hat{R}R}{2}+\frac{\hat{r}r}{2}+\frac{\hat{Q}Q}{2}+\frac{\hat{q}q}{2} \\
 & +\int Dz\log\int dw\exp\left[-\frac{\hat{R}+\hat{r}}{2}w^{2}+\left(\sqrt{\hat{r}}z-\frac{\lambda\beta M}{\sqrt{N}}\textrm{ sign}\left(w\right)\right)w\right].
\end{align*}

For $\beta\to\infty$, according to the characteristic of the Boltzmann distribution, the following scaling relations are assumed to hold,
i.e., 
\begin{equation}
\begin{cases}
\hat{Q}+\hat{q} & \equiv \beta E\\
\hat{q} & \equiv\beta^{2}F\\
\hat{R}+\hat{r} & \equiv \beta K\\
\hat{r} & \equiv \beta^{2}H\\
\beta\left(Q-q\right) & \equiv \chi\\
\beta\left(R-r\right) & \equiv \eta
\end{cases}
\end{equation}

Finally,  the entropy term is computed as 
\begin{align}
S&=\left(-ER+F\eta\right)G^{'}\left(-E\eta\right)+\frac{1}{2}EQ-\frac{1}{2}F\chi+\frac{1}{2}KR-\frac{1}{2}H\eta- \\
&\int\underset{w}{\min}\left\{ \frac{K}{2}w^{2}-\left(\sqrt{H}z-\frac{\lambda M}{\sqrt{N}}\textrm{ sign}\left(w\right)\right)w\right\} Dz.\label{eq:S-term-result}
\end{align}

\subsection{\label{Append-sub-Free energy density result} Free energy density result}

Combining the results (\ref{eq:Kesi-result}) and (\ref{eq:S-term-result}) together,
the free energy density for general loss function $\ell\left(\cdot\right)$
in the limit $\beta\rightarrow\infty$ is obtained as
\begin{align}
f\left(\beta\rightarrow\infty\right) & =-\underset{\mathtt{\varTheta}}{\textrm{Extr}}\left\{ \begin{array}{c}
-\alpha \mathbb{E}_{s,z} \left( \underset{y}{\min}\left[\frac{\left(y-s_{0}\left(\sqrt{Q}z+\sum_{j\in\Psi}\text{\ensuremath{\bar{J}_{j}}}s_{j}\right)\right)^{2}}{2\chi}+\ell\left(y\right)\right]\right)-\alpha\lambda\sum_{j\in\Psi}\left|\bar{J}_{j}\right|\\
+\left(-ER+F\eta\right)G^{'}\left(-E\eta\right)+\frac{1}{2}EQ-\frac{1}{2}F\chi\\
+\frac{1}{2}KR-\frac{1}{2}H\eta-\mathbb{E}_{z} \left(\underset{w}{\min}\left\{ \frac{K}{2}w^{2}-\left(\sqrt{H}z-\frac{\lambda M}{\sqrt{N}}\textrm{ sign}\left(w\right)\right)w\right\}\right)
\end{array}\right\} ,\label{eq:free-energy-result-general}
\end{align}
where the values of the parameters $\varTheta=\left\{ \chi,Q,E,R,F,\eta,K,H,\text{\ensuremath{\left\{  \bar{J}_{j}\right\} } }_{j\in\Psi}\right\} $
can be calculated by the extremization condition, i.e., solving the equations of state (EOS). For general loss
function $\ell\left(y\right)$, the EOS for (\ref{eq:free-energy-result-general}) is as follows
\begin{equation}
\begin{cases}
\hat{y}\left(s,z,\chi,Q,J\right)=\underset{y}{\arg\max}\left\{ -\frac{\left(y-s_{0}\left(\sqrt{Q}z+\sum_{j\in\Psi}\text{\ensuremath{\bar{J}_{j}}}s_{j}\right)\right)^{2}}{2\chi}-\ell\left(y\right)\right\} \\
E=\frac{\alpha}{\sqrt{Q}} \mathbb{E}_{s,z}\left( s_{0}z\frac{d\ell\left(y\right)}{dy}\mid_{y=\hat{y}\left(s,z,\chi,Q,J\right)}\right)\\
F=\alpha\mathbb{E}_{s,z}\left( \left(\frac{d\ell\left(y\right)}{dy}\mid_{y=\hat{y}\left(s,z,\chi,Q,J\right)}\right)^{2}\right)\\
R=\frac{1}{K^{2}}\left[\left(H+\frac{\lambda^{2}M^{2}}{N}\right)\textrm{erfc}\left(\frac{\lambda M}{\sqrt{2HN}}\right)-2\lambda M\sqrt{\frac{H}{N}}\frac{1}{\sqrt{2\pi}}e^{-\frac{\lambda^{2}M^{2}}{2HN}}\right]\\
E\eta=-\int\frac{\rho\left(\gamma\right)}{\tilde{\varLambda}-\gamma}d\gamma\\
Q=\frac{F}{E^{2}}+R\tilde{\varLambda}-\left(-ER+F\eta\right)\eta\frac{1}{\int\frac{\rho\left(\lambda\right)}{\left(\tilde{\varLambda}-\lambda\right)^{2}}d\lambda}\\
K=E\tilde{\varLambda}+\frac{1}{\eta}\\
\chi=\frac{1}{E}+\eta\tilde{\varLambda}\\
H=\frac{R}{\eta^{2}}+F\tilde{\varLambda}+\left(-ER+F\eta\right)E\frac{1}{\int\frac{\rho\left(\lambda\right)}{\left(\tilde{\varLambda}-\lambda\right)^{2}}d\lambda}\\
\eta=\frac{1}{K}\textrm{erfc}\left(\frac{\lambda M}{\sqrt{2HN}}\right)\\
\bar{J}_{j,j\in\Psi}=\underset{{J}_{j,j\in\Psi}}{\arg\min} \; \left\{\mathbb{E}_{s,z} \left( \left[\frac{\left(\hat{y}\left(s,z,\chi,Q,J\right)-s_{0}\left(\sqrt{Q}z+\sum_{j\in\Psi}\text{\ensuremath{{J}_{j}}}s_{j}\right)\right)^{2}}{2\chi}+\ell\left(\hat{y}\left(s,z,\chi,Q,J\right)\right)\right]\right)+\lambda\sum_{j\in\Psi}\left|{J}_{j}\right| \right\}
\end{cases}\label{eq:EOS-genereal-result}
\end{equation}
where $\tilde{\varLambda}$ satisfying $E\eta=-\int\frac{\rho\left(\gamma\right)}{\tilde{\varLambda}-\gamma}d\gamma $ is determined by the extremization condition in (\ref{eq:Gx-def}) combined with the free energy result (\ref{eq:free-energy-result-general}). In general, there are no analytic solutions for the EOS (\ref{eq:EOS-genereal-result}) but it can be solved numerically.

\subsubsection{quadratic loss $\ell\left(y\right)=\left(y-1\right)^{2}/2$}

In the case of square lass $\ell\left(y\right)=\left(y-1\right)^{2}/2$ for the $\ell_1$-LinR estimator,
there is an analytic solution to $y$ in $\underset{y}{\min}\left[\frac{\left(y-s_{0}\left(\sqrt{Q}z+\sum_{j\in\Psi}\text{\ensuremath{\bar{J}_{j}}}s_{j}\right)\right)^{2}}{2\chi}+\ell\left(y\right)\right]$ and thus the results can be further simplified. Specifically, the free energy can be written as follows
\begin{align}
f\left(\beta\rightarrow\infty\right) & =-\underset{\mathtt{\varTheta}}{\textrm{Extr}}\left\{\begin{array}{c}
-\frac{\alpha}{2\left(1+\chi\right)}\mathbb{E}_{s,z}\left[ \left(s_{0}-\sum_{j\in\Psi}s_{j}\bar{J}_{j}-\sqrt{Q}z\right)^{2}\right]-\alpha\lambda\sum_{j\in\Psi}\left|\bar{J}_{j}\right|\\
+\left(-ER+F\eta\right)G^{'}\left(-E\eta\right)+\frac{1}{2}EQ-\frac{1}{2}F\chi\\
+\frac{1}{2}KR-\frac{1}{2}H\eta-\mathbb{E}_{z}\left[\underset{w}{\min}\left\{ \frac{K}{2}w^{2}-\left(\sqrt{H}z-\frac{\lambda M}{\sqrt{N}}\textrm{ sign}\left(w\right)\right)w\right\}\right]
\end{array}\right\} ,\label{eq:free-energy-result-linear}
\end{align}
and the corresponding EOS
can be written as 
\begin{equation}
\begin{cases}
E=\frac{\alpha}{\left(1+\chi\right)}, & \left(a\right)\\
F=\frac{\alpha}{\left(1+\chi\right)^{2}}\left[\mathbb{E}_{s} \left(s_{i}-\sum_{j\in\Psi}s_{j}\bar{J}_{j}\right)^{2}+Q\right], & \left(b\right)\\
R=\frac{1}{K^{2}}\left[\left(H+\frac{\lambda^{2}M^{2}}{N}\right)\textrm{erfc}\left(\frac{\lambda M}{\sqrt{2HN}}\right)-2\lambda M\sqrt{\frac{H}{N}}\frac{1}{\sqrt{2\pi}}e^{-\frac{\lambda^{2}M^{2}}{2HN}}\right], & \left(c\right)\\
E\eta=-\int\frac{\rho\left(\gamma\right)}{\tilde{\varLambda}-\gamma}d\gamma, & \left(d\right)\\
Q=\frac{F}{E^{2}}+R\tilde{\varLambda}-\left(-ER+F\eta\right)\frac{\eta}{\int\frac{\rho\left(\gamma\right)}{\left(\tilde{\varLambda}-\gamma\right)^{2}}d\gamma}, & \left(e\right)\\
K=E\tilde{\varLambda}+\frac{1}{\eta}, & \left(f\right)\\
\chi=\frac{1}{E}+\eta\tilde{\varLambda}, & \left(g\right)\\
H=\frac{R}{\eta^{2}}+F\tilde{\varLambda}+\left(-ER+F\eta\right)\frac{E}{\int\frac{\rho\left(\gamma\right)}{\left(\tilde{\varLambda}-\gamma\right)^{2}}d\gamma}, & \left(h\right)\\
\eta=\frac{1}{K}\textrm{erfc}\left(\frac{\lambda M}{\sqrt{2HN}}\right), & \left(i\right)\\
\bar{J}_{j}=\frac{\mathtt{soft}\left(\tanh\left(K_{0}\right),\lambda\left(1+\chi\right)\right)}{1+\left(d-1\right)\tanh^{2}\left(K_{0}\right)},j\in\Psi, & \left(j\right)
\end{cases}\label{eq:fixed-point-equations-linear-meanJ}
\end{equation}
Note that the mean estimates $\left\{ \bar{J}_{j},j\in\Psi\right\} $ in (\ref{eq:fixed-point-equations-linear-meanJ}) is obtained by solving the following reduced optimization problem
\begin{equation}
\underset{\left\{ \bar{J}_{j}\right\}}{\arg\min}\left\{\frac{1}{2\left(1+\chi\right)}\mathbb{E}_{s,z}\left[ \left(s_{0}-\sum_{j\in\Psi}s_{j}\bar{J}_{j}-\sqrt{Q}z\right)^{2}\right]-\lambda\sum_{j\in\Psi}\left|\bar{J}_{j}\right|\right\}, \label{eq:J-mean-result-EOS}
\end{equation}
where the corresponding fixed-point equation associated with any $\bar{J}_{k},k\in\Psi$ can be written
as follows
\begin{equation}
\frac{1}{1+\chi}\mathbb{E}_{s}\left[ s_{k}\left(s_{0}-\sum_{j\in\Psi}s_{j}\bar{J}_{j}\right)\right]-\lambda\mathtt{sign}\left(\bar{J}_{k}\right)=0,\forall k\in\Psi, \label{eq:EOS-J-mean-fixed-point}
\end{equation}
where the $\mathtt{sign}(\cdot)$ denotes an element-wise application of the standard sign function.  For a RR graph $G\in \mathcal{G}_{N,d,K_{0}}$ with degree $d$
and coupling strength $K_{0}$, without loss of generality, assuming that all the active couplings are positive,  we have $\mathbb{E}_{s}\left(s_0 s_k\right) = \tanh\left({K_0}\right), \forall k \in \Psi$, and $\mathbb{E}_{s}\left(s_k s_j\right) = \tanh^2\left({K_0}\right),\; \forall k,j \in \Psi, k\neq j$. Given these results and thanks to the the symmetry, we obtain 
\begin{equation}
\bar{J}_{j}=\frac{\mathtt{soft}\left(\tanh\left(K_{0}\right),\lambda\left(1+\chi\right)\right)}{1+\left(d-1\right)\tanh^{2}\left(K_{0}\right)},j\in\Psi,  \label{eq:EOS-J-mean-solution}
\end{equation}
where $\mathtt{soft}\left(z,\tau\right)=\mathtt{sign}\left(z\right)\left(\left|z\right|-\tau\right)_{+}$
is the soft-thresholding function, i.e., 
\begin{align}
\mathtt{soft}\left(z,\tau\right)\equiv \mathtt{sign}\left(z\right)\left(\left|z\right|-\tau\right)_{+} \equiv\begin{cases}
z-\tau, & z>\tau\\
0, & \left|z\right|\leq\tau\\
z+\tau, & z<-\tau
\end{cases}
\end{align}
On the other hand, in the inactive set $\bar{\Psi}$, each component of the scaled noise estimates can be statistically described as the solution to the scalar estimator $\underset{w}{\min}\left\{ \frac{K}{2}w^{2}-\left(\sqrt{H}z-\frac{\lambda M}{\sqrt{N}}\textrm{ sign}\left(w\right)\right)w\right\} $
in (\ref{eq:free-energy-result-general}). Consequently, recalling the definition of $w$ in (\ref{eq:sparse-anastz}), the
estimates $\left\{ \hat{J}_{j},j\in\bar{\Psi}\right\} $ in the inactive set $\bar{\Psi}$ are
\begin{align}
\hat{J}_{j} & =\frac{\sqrt{H}}{K\sqrt{N}}\mathtt{soft}\left(z_{j},\frac{\lambda M}{\sqrt{HN}}\right)\nonumber \\
 & =\underset{J_{j}}{\arg\min}\left[\frac{1}{2}\left(J_{j}-\frac{1}{K}\sqrt{\frac{H}{N}}z_{j}\right)^{2}+\frac{\lambda M}{KN}\left|J_{j}\right|\right], j\in\bar{\Psi}, \label{eq:J-inactive-decoupled-result}
\end{align}
which $z_j\sim\mathcal{N}\left(0,1\right),j\in\bar{\Psi}$ are i.i.d. random Gaussian noise. 

Consequently, it can be
seen that from (\ref{eq:EOS-J-mean-solution}) and (\ref{eq:J-inactive-decoupled-result}), statistically, the $\ell_1$-LinR estimator is  decoupled into two scalar thresholding estimators for the active set $\Psi$ and inactive set $\bar{\Psi}$, respectively.

\subsubsection{Logistic loss $\ell\left(y\right)=\log\left(1+e^{-2y}\right)$}

In the case of logistic lass $\ell\left(y\right)=\log\left(1+e^{-2y}\right)$ for the $\ell_1$-LogR estimator,
however, there is no analytic solution to $y$ in $\underset{y}{\min}\left[\frac{\left(y-s_{0}\left(\sqrt{Q}z+\sum_{j\in\Psi}\text{\ensuremath{\bar{J}_{j}}}s_{j}\right)\right)^{2}}{2\chi}+\ell\left(y\right)\right]$
and we have to solve it together iteratively with other parameters $\varTheta$. After some algebra, we obtain the EOS for the $\ell_1$-LogR estimator: 
\begin{equation}
\begin{cases}
\frac{\hat{y}\left(s,z,\chi,Q,J\right)-s_{0}\left(\sqrt{Q}z+\sum_{j\in\Psi}\text{\ensuremath{\bar{J}_{j}}}s_{j}\right)}{\chi} = 1 - \tanh\left(\hat{y}\left(s,z,\chi,Q,J\right)\right),\\
E=\alpha\mathbb{E}_{s,z}\left( \frac{s_{0}z}{\sqrt{Q}}\tanh\left(\hat{y}\left(S,z,\chi,Q,J\right)\right)\right),\\
F=\alpha\mathbb{E}_{s,z}\left( \left(1-\tanh\left(\hat{y}\left(S,z,\chi,Q,J\right)\right)\right)^{2}\right),\\
R=\frac{1}{K^{2}}\left[\left(H+\frac{\lambda^{2}M^{2}}{N}\right)\textrm{erfc}\left(\frac{\lambda M}{\sqrt{2HN}}\right)-2\lambda M\sqrt{\frac{H}{N}}\frac{1}{\sqrt{2\pi}}e^{-\frac{\lambda^{2}M^{2}}{2HN}}\right],\\
E\eta=-\int\frac{\rho\left(\gamma\right)}{\tilde{\varLambda}-\gamma}d\gamma,\\
Q=\frac{F}{E^{2}}+R\tilde{\varLambda}-\left(-ER+F\eta\right)\eta\frac{1}{\int\frac{\rho\left(\lambda\right)}{\left(\tilde{\varLambda}-\lambda\right)^{2}}d\lambda},\\
K=E\tilde{\varLambda}+\frac{1}{\eta},\\
\chi=\frac{1}{E}+\eta\tilde{\varLambda},\\
H=\frac{R}{\eta^{2}}+F\tilde{\varLambda}+\left(-ER+F\eta\right)E\frac{1}{\int\frac{\rho\left(\lambda\right)}{\left(\tilde{\varLambda}-\lambda\right)^{2}}d\lambda},\\
\eta=\frac{1}{K}\textrm{erfc}\left(\frac{\lambda M}{\sqrt{2HN}}\right),\\
\bar{J}_{j}=\frac{\mathtt{soft}\left(\mathbb{E}_{s,z}\left( \hat{y}\left(s,z,\chi,Q,J\right)s_{0}\sum_{j\in\Psi}s_{j}\right),\lambda d\chi\right)}{d\left(1+\left(d-1\right)\tanh^{2}\left(K_{0}\right)\right)},j\in\Psi.
\end{cases}\label{eq:fixed-point-equations-logistic-meanJ}
\end{equation}
In the active set $\Psi$, the mean estimates $\left\{ \bar{J}_{j},j\in\Psi\right\} $
can be obtained by solving a reduced $\ell_{1}$-regularized optimization
problem
\begin{equation}
\underset{\left\{ \bar{J}_{j}\right\} _{j\in\Psi}}{\min}\left\{ \mathbb{E}_{s,z} \left( \underset{y}{\min}\left[\frac{\left(y-s_{0}\left(\sqrt{Q}z+\sum_{j\in\Psi}\text{\ensuremath{\bar{J}_{j}}}s_{j}\right)\right)^{2}}{2\chi}+\log\left(1+e^{-2y}\right)\right]\right)+\lambda\sum_{j\in\Psi}\left|\bar{J}_{j}\right|\right\} .\label{eq:J-mean-decoupled-result}
\end{equation}
In contrast to  the $\ell_1$-LinR estimator, the mean estimates  $\left\{ \bar{J}_{j},j\in\Psi\right\}$ in (\ref{eq:J-mean-decoupled-result}) for the $\ell_1$-LogR estimator do not have analytic solutions and also have to be solved numerically. For a RR graph $G\in \mathcal{G}_{N,d,K_{0}}$ with degree $d$
and coupling strength $K_{0}$, after some algebra, the corresponding fixed-point equations for $\left\{ \bar{J}_{j}=J,j\in\Psi\right\}$ are obtained as follows
\begin{align}
J & =\frac{\mathtt{soft}\left(\mathbb{E}_{s,z}\left( \hat{y}\left(s,z,\chi,Q,J\right)s_{0}\sum_{j\in\Psi}s_{j}\right),\lambda d\chi\right)}{d\left(1+\left(d-1\right)\tanh^{2}\left(K_{0}\right)\right)},\label{eq:J-result-mean-LR}
\end{align}
which can be solved iteratively. 

The estimates in the inactive set $\bar{\Psi}$ are the same as (\ref{eq:J-inactive-decoupled-result}) that of $\ell_1$-LinR, which can be described by a scalar theresholding estimator once the EOS is solved. 

\section{\label{sec:Verification-of-the}Check the consistency of ansatz (A1) }

\begin{figure}
\centering{}\includegraphics[width=10cm]{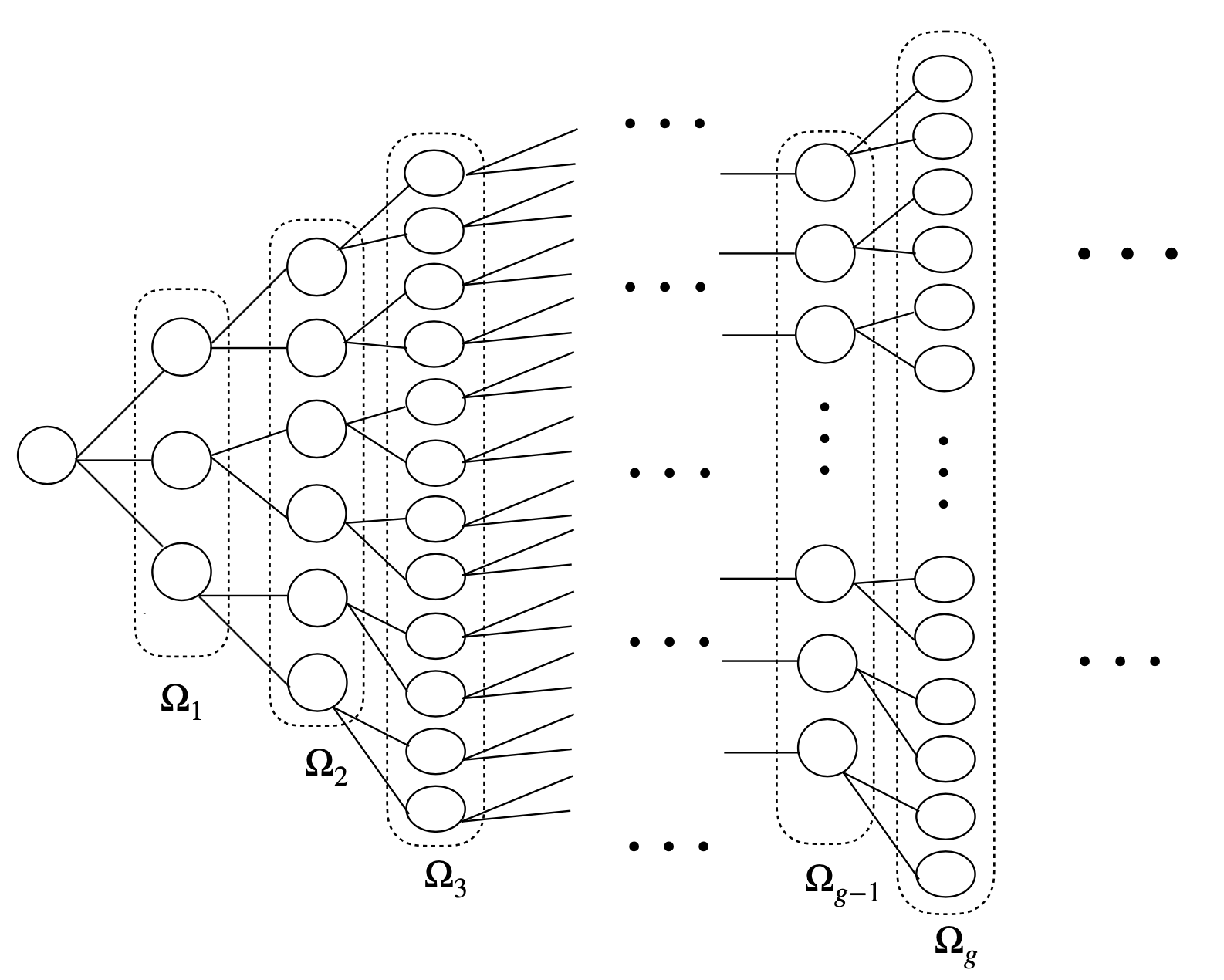}\caption{Schematic of generations of spins. In general, the $g$-th generation
of spin $s_{0}$ is denoted as $\Omega_{g}$, whose distance from spin
$s_{0}$ is $g$. \label{fig:tree-graph}}
\end{figure}

To check the consistency of Ansatz (A1), first we categorize the estimators based
on the distance or generation from the focused spin $s_{0}$. Considering
the original Ising model whose coupling network is a tree-like graph,
we can naturally define generations of the spins according to the
distance from the focused spin $s_{0}$. We categorize the spins directly
connected to $s_{0}$ as the first generation and denote the corresponding
index set as $\Omega_{1}=\{i|J_{i}^{*}\neq0,i\in\left\{ 1,\ldots,N-1\right\} \}$.
Each spin in $\Omega_{1}$ is connected to some other spins except
for $s_{0}$, and those spins constitute the second generation and
we denote its index set as $\Omega_{2}$. This recursive construction
of generations can be unambiguously continued on the tree-like graph,
and we denote the index set of the $g$-th generation from spin $s_{0}$
as $\Omega_{g}$. The overall construction of generations is graphically
represented in Fig. \ref{fig:tree-graph}. 
Generally, assume that the set of nonzero values of the $\ell_{1}$-LinR estimator is denoted as $\Psi=\left\{ \Omega_{1},\ldots,\Omega_{g}\right\} $.
Then, Ansatz (A1) means that the correct active set of the mean estimates is $\Psi=\left\{ \Omega_{1}\right\} $. 

To verify this, we examine the values of mean estimates based on (\ref{eq:free-energy-result-linear}). Due to the symmetry, it is expected that for each $a=1,...,g$, the
values of the mean estimates $\bar{J}_{j\in\Omega_{a}}=J_{a}$
are identical to each other within the same set $\Omega_{a},a=1...g$.
In addition, if the solutions satisfy Ansatz (A1) in (\ref{eq:sparse-anastz}),
i.e., $J_{1}=J,J_{a}=0,a\geq2$, from  (\ref{eq:free-energy-result-linear}) we obtain 
\begin{equation}
\begin{cases}
\frac{1}{1+\chi}\left[\tanh\left(K_{0}\right)-\left(1+\left(d-1\right)\tanh^{2}\left(K_{0}\right)\right)J\right]-\lambda=0, & j\in\Omega_{1};\\
\left|\frac{1}{1+\chi}\left[\tanh^{a}\left(K_{0}\right)-\tanh^{a-1}\left(K_{0}\right)\left(1+\left(d-1\right)\tanh^{2}\left(K_{0}\right)\right)J\right]\right|\leq\lambda, & j\in\Omega_{a},a\geq2,
\end{cases}\label{eq:EOS-differentNN}
\end{equation}
where the result $\mathbb{E}_{s}\left(s_i s_j\right) = \tanh^{d_0}\left({K_0}\right)$ is used for any two spins $s_i,s_j$ whose distance is $d_0$ in the RR graph $G\in \mathcal{G}_{N,d,K_{0}}$.
Note that the solution of the first equation in (\ref{eq:EOS-differentNN})
automatically satisfies the second equation (sub-gradient condition)
since $\left|\tanh\left(K_{0}\right)\right|\leq1$, which indicates
that $J_{1}=J,J_{a}=0,a\geq2$ is one valid solution. Moreover, the convexity
of the quadratic loss function indicates that this is the unique and
correct solution, which checks the Ansatz (A1). 

\section{\label{sec:Haar-Orthogonal-Assumption}Check the consistency of ansatz (A2) }

\begin{figure}
\includegraphics[width=15cm]{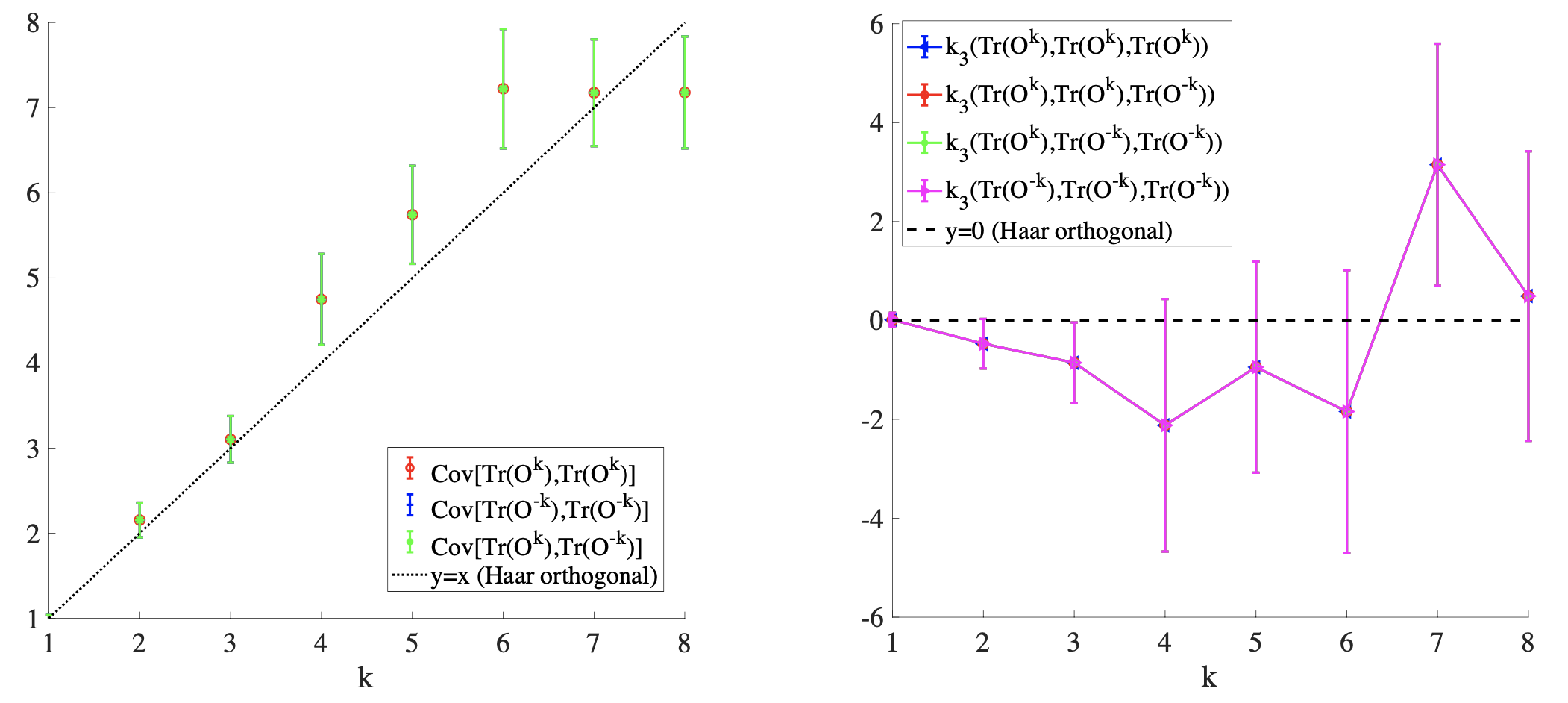}

\caption{The RR graph $G\in \mathcal{G}_{N,d,K_{0}}$ with  $N=1000,d=3,K_{0}=0.4$ is generated and we  
compute the associated covariance matrix $C$ and then diagonalize it as $C=O\Lambda O^{T}$, obtaining the orthogonal matrix $O$. Then
the $\mathtt{Tr}\left(O^{k}\right),\mathtt{Tr}\left(O^{-k}\right)$ for several $k$ ($k=1\sim8$) are computed, where $\mathtt{Tr}\left(\cdot\right)$ is the trace operation. This procedure is repeated 200 times with different random numbers, from which we obtain the ensemble of $\mathtt{Tr}\left(O^{k}\right)$ and $\mathtt{Tr}\left(O^{-k}\right)$.
Consequently, the cumulants of 1st, 2nd, and 3rd orders are computed. All of them exhibit the expected theoretical behavior.\label{fig:Haar-support}}
\end{figure}

We here check the consistency of a part of the Ansatz (A2) in Sec.\ref{subsec:Free-energy-calculation}, the orthogonal matrix $O$ diagonalizing the covariance matrix $C$ is distributed from the Haar orthogonal measure. To achieve
this, we compare certain properties of the orthogonal  matrix generated
from the diagonalization of the covariance matrix $C$ with the orthogonal
 matrix which is actually generated from the Haar orthogonal measure.
Specifically, we compute the cumulants of the trace of the power $k$
of the orthogonal  matrix. All cumulants with degree
$r\geq3$ are shown to disappear in the large $N$ limit \citep{diaconis1994eigenvalues,johansson1997random}. The nontrivial cumulants
are only second order cumulant with the same power $k$. We have computed
these cumulants about the orthogonal  matrix from the covariance matrix $C$ and
found that they exhibit the same behavior as the ones generated from
the true Haar measure, as shown in Fig. \ref{fig:Haar-support}. 

\section{\label{sec:Perfect-graph-recovery}Details of the High-dimensional asymptotic result}

Here the asymptotic performance of  $Precision$ and $Recall$ are considered for both the $\ell_1$-LinR estimator and the $\ell_1$-LogR estimator. Recall that perfect Ising model selection is achieved if and only if $Precision=1$ and $Recall=1$ 
\subsection{\label{subsec:Recall-rate}Recall rate}

According to the definition in (\ref{eq:Recall-def-1}), the recall rate is only related to the statistical properties of estimates in the active set $\Psi$ and thus the mean estimates $\left\{ \bar{J}_{j}\right\} _{j\in\Psi}$ in the limit  $M\to \infty$ are considered. 

\subsubsection{quadratic loss}

In this case, in the limit $M\rightarrow\infty$, the mean estimates $\left\{ \bar{J}_{j}=J\right\} _{j\in\Psi}$ in the active set $\Psi$ are shown in (\ref{eq:EOS-J-mean-solution}) and rewritten as follows for ease of reference
\begin{equation}
J=\frac{\mathtt{soft}\left(\tanh\left(K_{0}\right),\lambda\left(1+\chi\right)\right)}{1+\left(d-1\right)\tanh^{2}\left(K_{0}\right)}. \label{eq:J-result-soft}
\end{equation}
As a result, as long as $\lambda\left(1+\chi\right)<\tanh\left(K_{0}\right)$,
$J>0$ and thus we can successfully recover 
the active set so that $Recall=1$. In addition, when $M=\mathcal{O}\left(\log N\right)$,
$\chi\rightarrow0$ as $N\rightarrow\infty$, as demonstrated later by the relation  in (\ref{eq:chi-result}). As a result, the regularization
parameter needs to satisfy $0<\lambda<\tanh\left(K_{0}\right)$. 

\subsubsection{Logistic loss }
In this case, in the limit $M\rightarrow\infty$, the mean estimates $\left\{ \bar{J}_{j}=J\right\} _{j\in\Psi}$ in the active set $\Psi$ are shown in (\ref{eq:J-result-mean-LR}) and rewritten as follows for ease of reference
\begin{align}
J & =\frac{\mathtt{soft}\left(\mathbb{E}_{s,z}\left( \hat{y}\left(s,z,\chi,Q,J\right)s_{0}\sum_{j\in\Psi}s_{j}\right),\lambda d\chi\right)}{d\left(1+\left(d-1\right)\tanh^{2}\left(K_{0}\right)\right)}.\label{eq:J-result-mean-LR-2}
\end{align}
There is no analytic solution for $\hat{y}\left(s,z,\chi,Q,J\right)$
and the following fixed-point equation has to be solved numerically
\begin{align}
\frac{\hat{y}\left(s,z,\chi,Q,J\right)-s_{0}\left(\sqrt{Q}z+J\sum_{j\in\Psi}s_{j}\right)}{\chi} = 1 - \tanh\left(\hat{y}\left(s,z,\chi,Q,J\right)\right).  \label{eq:J-mean-LR-fixed-point}
\end{align}
Then one can determine the valid choice of $\lambda$ to enable $J>0$. 
Numerical results show that the choice of $\lambda$ is similar to that of the quadratic loss. 

\subsection{Precision rate}

According to the definition in (\ref{eq:Recall-def-1}), to compute the $Precision$, the number of true positives $TP$ and false positives $FP$ are needed, respectively. On the one hand, as discussed in Appendix \ref{subsec:Recall-rate}, in the limit $M\to \infty$, the recall rate approach to one and thus we have $TP=d$ for a RR graph $G\in \mathcal{G}_{N,d,K_{0}}$. On the other hand, the number of false positives $FP$ can be computed as $FP=FPR\cdot N$, where $FPR$ is the false positive rate (FPR). 

As shown in Appendix
\ref{Append-sub-Free energy density result}, the estimator in the inactive
set $\bar{\Psi}$ can be statistically described by a scalar estimator (\ref{eq:J-inactive-decoupled-result}) and thus the $FPR$
can be computed as
\begin{equation}
FPR=\textrm{erfc}\left(\frac{\lambda M}{\sqrt{2HN}}\right),\label{eq:FPR-result}
\end{equation}
which depends on $\lambda,M,N,H$. However, for both the quadratic loss and logistic loss, there is no analytic result for $H$ in (\ref{eq:EOS-genereal-result}). Nevertheless, we can obtain some asymptotic result
using perturbative analysis. 

Specifically, we focus on the asymptotic behavior of the macroscopic parameters,
e.g., $\chi,Q,K,E,H,F$, in the regime $FPR\rightarrow0$, which is necessary for successful Ising model selection. 
From $\eta=\frac{1}{K}\textrm{erfc}\left(\frac{\lambda M}{\sqrt{2HN}}\right)$ in EOS (\ref{eq:EOS-genereal-result}) and the $FPR$ in (\ref{eq:FPR-result}), there is $FPR=K\eta$. Moreover, by combining $E\eta=-\int\frac{\rho\left(\gamma\right)}{\tilde{\varLambda}-\gamma}d\gamma$ and $K=E\tilde{\varLambda}+\frac{1}{\eta}$, the following relation  can be obtained 
\begin{equation}
\textrm{erfc}\left(\frac{\lambda M}{\sqrt{2HN}}\right)=1-\int\frac{\rho\left(\gamma\right)}{1-\frac{\gamma}{\tilde{\varLambda}}}d\gamma.\label{eq:Lamda-relationship}
\end{equation}
Thus as $FPR=\textrm{erfc}\left(\frac{\lambda M}{\sqrt{2HN}}\right)\rightarrow0$,
there is $\int\frac{\rho\left(\gamma\right)}{1-\frac{\gamma}{\tilde{\varLambda}}}d\gamma\rightarrow1$, implying that the magnitude of $\tilde{\varLambda}\to \infty$. Consequently, using the truncated series expansion, we obtain
\begin{align}
E\eta & =-\int\frac{\rho\left(\gamma\right)}{\tilde{\varLambda}-\gamma}d\gamma\nonumber \\
 & =-\frac{1}{\tilde{\varLambda}}\sum_{k=0}^{\infty}\frac{\left\langle \gamma^{k}\right\rangle }{\tilde{\varLambda}^{k}}\nonumber \\
 & \simeq -\frac{1}{\tilde{\varLambda}}-\frac{\left\langle \gamma\right\rangle }{\tilde{\varLambda}^{2}},\label{eq:Lambda-approx}
\end{align}
where $\left\langle \gamma^{k}\right\rangle =\int\rho\left(\gamma\right)\gamma^{k}d\gamma$.
Then, solving the quadratic equation (\ref{eq:Lambda-approx}), we
obtain the solution (the other solution is not considered since it
is a smaller value) of $\tilde{\varLambda}$ as
\begin{align}
\tilde{\varLambda} & =\frac{-1-\sqrt{1-4E\eta\left\langle \gamma\right\rangle }}{2E\eta}\simeq \left\langle \gamma\right\rangle -\frac{1}{E\eta}.\label{eq:Lambda-solution}
\end{align}
To compute $\int\frac{\rho\left(\gamma\right)}{\left(\tilde{\varLambda}-\gamma\right)^{2}}d\gamma$, we use the following relation  
\begin{align}
f\left(\tilde{\varLambda}\right) & =-\int\frac{\rho\left(\gamma\right)}{\tilde{\varLambda}-\gamma}d\gamma\simeq-\frac{1}{\tilde{\varLambda}}-\frac{\left\langle \gamma\right\rangle }{\tilde{\varLambda}^{2}},\label{eq:Taylor-result-1}\\
\frac{df\left(\tilde{\varLambda}\right)}{d\tilde{\varLambda}} & =\int\frac{\rho\left(\gamma\right)}{\left(\tilde{\varLambda}-\gamma\right)^{2}}d\gamma\simeq\frac{1}{\tilde{\varLambda}^{2}}+2\frac{\left\langle \gamma\right\rangle }{\tilde{\varLambda}^{3}}.\label{eq:Taylor-result-2}
\end{align}

Substituting the results (\ref{eq:Lambda-solution}) - (\ref{eq:Taylor-result-2})
into (\ref{eq:EOS-genereal-result}), after some algebra, we obtain
\begin{align}
K & \simeq E\left\langle \gamma\right\rangle ,\label{eq:K-result}\\
\chi & \simeq \eta\left\langle \gamma\right\rangle ,\label{eq:chi-result}\\
Q & \simeq \frac{\left\langle \gamma\right\rangle ^{3}E^{2}\eta^{2}R-\left\langle \gamma\right\rangle ^{3}EF\eta^{3}+3\left\langle \gamma\right\rangle ^{2}F\eta^{2}-R\left\langle \gamma\right\rangle }{3E\eta\left\langle \gamma\right\rangle -1},\label{eq:Q-result}\\
H & \simeq \frac{\left\langle \gamma\right\rangle ^{3}E^{2}\eta^{2}F-\left\langle \gamma\right\rangle ^{3}R\eta E^{3}+3\left\langle \gamma\right\rangle ^{2}RE^{2}-F\left\langle \gamma\right\rangle }{3E\eta\left\langle \gamma\right\rangle -1}.\label{eq:H-result}
\end{align}

\begin{figure}
\includegraphics[width=16cm]{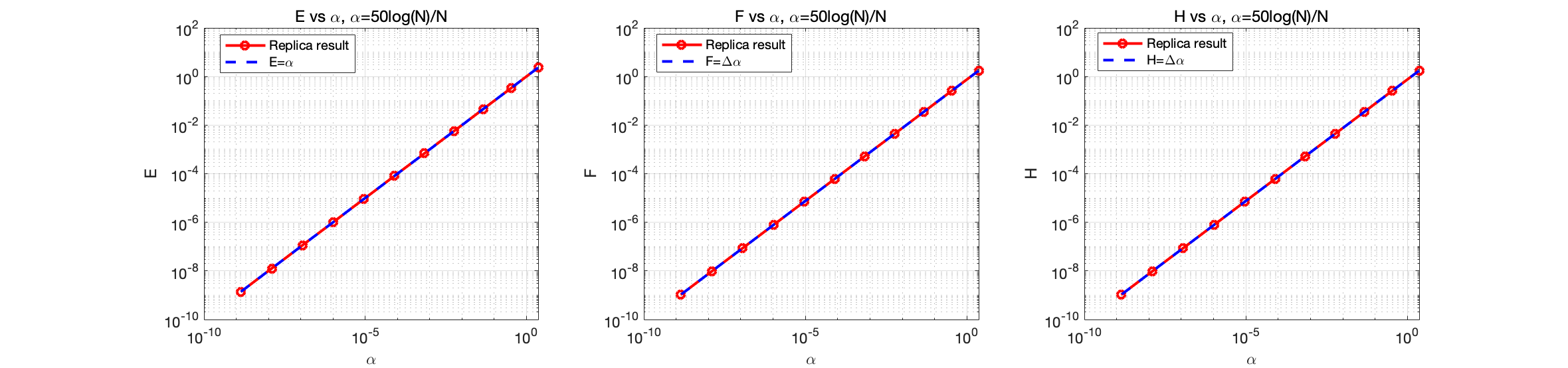}\caption{$E,F,H$ versus $\alpha$ when $\alpha=50(\log N)/N$ for $N=10^{2}\sim10^{12}$ for RR graph $G\in \mathcal{G}_{N,d,K_{0}}$  with $d=3,K_0=0.4$. Note that in this case,
there is $\left\langle \gamma\right\rangle =1$. \label{fig:EFH_vs_alpha}}
\end{figure}

In addition, as $FPR=\textrm{erfc}\left(\frac{\lambda M}{\sqrt{2HN}}\right)\rightarrow0$, from (\ref{eq:EOS-genereal-result}) we obtain 
\begin{align}
R & =\frac{1}{K^{2}}\left[\left(H+\frac{\lambda^{2}M^{2}}{N}\right)\textrm{erfc}\left(\frac{\lambda M}{\sqrt{2HN}}\right)-2\lambda M\sqrt{\frac{H}{N}}\frac{1}{\sqrt{2\pi}}e^{-\frac{\lambda^{2}M^{2}}{2HN}}\right]\nonumber \\
 & \simeq\frac{H}{K^{2}}\textrm{erfc}\left(\frac{\lambda M}{\sqrt{2HN}}\right)\simeq\frac{H}{K}\eta\simeq\frac{H}{E\left\langle \gamma\right\rangle}\eta,\label{eq:R-relation-HK}
\end{align}
where the first result in $\simeq$ uses the asymptotic relation  $\textrm{erfc}\left(x\right)\simeq\frac{1}{x\sqrt{\pi}}e^{-x^{2}}$ as $x \to \infty$ and the last result in $\simeq$ results from the asymptotic relation  in (\ref{eq:K-result}). Then, substituting (\ref{eq:R-relation-HK}) into (\ref{eq:H-result}) leads to the following relation 
\begin{align}
\left(3E\eta\left\langle \gamma\right\rangle -1\right)H	\simeq \left\langle \gamma\right\rangle ^{3}E^{2}\eta^{2}F-\left\langle \gamma\right\rangle ^{2}\eta^{2}E^{2}H+3E\eta\left\langle \gamma\right\rangle H-F\left\langle \gamma\right\rangle.  \label{eq:H-relation-simple}
\end{align}	
Interestingly, the common terms $3E\eta\left\langle \gamma\right\rangle H$ in both sides of (\ref{eq:H-relation-simple}) cancel with each other. Therefore, the key result for $H$ is obtained as follows 
\begin{align}
H \simeq F\left\langle \gamma\right\rangle. \label{eq:H-relation-F}
\end{align}
In addition, from (\ref{eq:H-relation-F}) and (\ref{eq:Q-result}), $Q$ can be simplified as  
\begin{align}
Q & \simeq R\left\langle \gamma\right\rangle. \label{eq:Q-relation}
\end{align}
As shown in (\ref{eq:EOS-genereal-result}), $F=\alpha\mathbb{E}_{s,z} \left(\frac{d\ell\left(y\right)}{dy}\mid_{y=\hat{y}\left(s,z,\chi,Q,J\right)}\right)^{2}$, thus the result $H\simeq F\left\langle \gamma\right\rangle$ in (\ref{eq:H-relation-F}) implies that there is a linear
relation between $H$ and $\alpha\equiv M/N$. The relation between
$E,F,H$ and $\alpha$ are also verified numerically in Fig. \ref{fig:EFH_vs_alpha}
when $M=50(\log N)$ for $N=10^{2}\sim10^{12}$ 
using the $\ell_{1}$-LinR estimator. 

In the paramagnetic phase, it can be obtained that the mean value of the eigenvalue $\left\langle \gamma\right\rangle$. Specifically, we have $\left\langle \gamma\right\rangle = \frac{1}{N}\sum_{i=1}^N \gamma_i  = \frac{1}{N} \textrm{Tr} C = (1/N) \times N = 1$.
Denote by $H\simeq F\left\langle \gamma\right\rangle \equiv \alpha\triangle$, where $\triangle=\mathbb{E}_{s,z} \left(\frac{d\ell\left(y\right)}{dy}\mid_{y=\hat{y}\left(s,z,\chi,Q,J\right)}\right)^{2}=\mathcal{O}\left(1\right)$, then the $FPR$ in (\ref{eq:FPR-result}) can be rewritten
as follows
\begin{align}
FPR & =\textrm{erfc}\left(\frac{\lambda M}{\sqrt{2\alpha\triangle N}}\right)\nonumber \\
 & =\textrm{erfc}\left(\lambda\sqrt{\frac{M}{2\triangle}}\right)\nonumber \\
 & \leq\frac{1}{\sqrt{\pi}}e^{-\frac{\lambda^{2}M}{2\triangle}-\frac{1}{2}\log\left(\frac{\lambda^{2}M}{2\triangle}\right)},\label{eq:FPR-result-upper-bound}
\end{align}
where the last inequality uses the upper bound of $\textrm{erfc}$
function, i.e., $\textrm{erfc}\left(x\right)\leq\frac{1}{x\sqrt{\pi}}e^{-x^{2}}$.
Consequently, the number of false positives $FP$ satisfies 
\begin{align}
FP & \leq\frac{N}{\sqrt{\pi}}e^{-\frac{\lambda^{2}M}{2\triangle}-\frac{1}{2}\log\left(\frac{\lambda^{2}M}{2\triangle}\right)}\nonumber \\
 & =\frac{1}{\sqrt{\pi}}e^{-\frac{\lambda^{2}M}{2\triangle}-\frac{1}{2}\log\left(\frac{\lambda^{2}M}{2\triangle}\right)+\log N}\nonumber \\
 & <\frac{1}{\sqrt{\pi}}e^{-\frac{\lambda^{2}M}{2\triangle}+\log N},\label{eq:FP-result}
\end{align}
where the last inequality holds when $\frac{\lambda^{2}M}{2\triangle}>1$,
which is necessary when $FP\rightarrow0$ as $N\rightarrow\infty$. Consequently, to ensure $FP\rightarrow0$ as $N\rightarrow\infty$, from (\ref{eq:FP-result}),
the term $\frac{\lambda^{2}M}{2\triangle}$ should grow at least faster than
$\log N$, i.e., 
\begin{align}
M & >\frac{2\triangle\log N}{\lambda^{2}}.\label{eq:M-condition}
\end{align}
Meanwhile, the number of false positives $FP$ will decay as $\mathcal{O}\left(e^{-c\log{N}}\right)$ for some constant $c \left(>0\right)$. 

\subsubsection{Quadratic loss}

In this case, when $0<\lambda<\tanh{\left(K_0\right)}$, from (\ref{eq:fixed-point-equations-linear-meanJ}),
we can obtain an analytic result for $\triangle$ as follows
\begin{align} 
\bigtriangleup & \simeq \mathbb{E}_{s_0} \left(s-\sum_{j\in\Psi}s_{j}\bar{J}_{j}\right)^{2} \\
 & =\frac{1-\tanh^{2}K_{0}+d\lambda^{2}}{1+\left(d-1\right)\tanh^{2}K_{0}}. \label{eq:scalings_results-1-1}
\end{align}
On the other hand, from the discussion in Appendix \ref{subsec:Recall-rate}, the recall rate $Recall\rightarrow1$ as $M\to \infty$  when $0<\lambda<\tanh K_{0}$. Overall, for
a RR graph $G\in \mathcal{G}_{N,d,K_{0}}$ with degree $d$ and coupling
strength $K_{0}$, given $M$ i.i.d. samples $\mathcal{D}^{M}=\left\{ \boldsymbol{s}^{\left(1\right)},...,\boldsymbol{s}^{\left(M\right)}\right\} $,
using $\ell_{1}$-LinR estimator (\ref{eq:PL-estimator-linear-def})
with regularization parameter $\lambda$, perfect recovery of the
graph structure $G$ can be achieved as $N\to\infty$ if the number of samples $M$
satisfies
\begin{align}
M & >\frac{c\left(\lambda,K_{0}\right)\log N}{\lambda^{2}},\lambda\in\left(0,\tanh\left(K_{0}\right)\right)\label{eq:C_threshold_lowerbound-1-1}
\end{align}
where $c\left(\lambda,K_{0}\right)$ is a value dependent on the regularization
parameter $\lambda$ and coupling strength $K_{0}$, which can be
approximated in the limit $N\rightarrow\infty$ as:
\begin{equation}
c\left(\lambda,K_{0}\right)=\frac{2\left(1-\tanh^{2}\left(K_{0}\right)+d\lambda^{2}\right) }{1+\left(d-1\right)\tanh^{2}\left(K_{0}\right)}.\label{eq:critical_c_value-1}
\end{equation}

\subsubsection{Logistic loss}

In this case, from (\ref{eq:fixed-point-equations-logistic-meanJ}),
the value of $\triangle$ can be computed as 
\begin{align}
\bigtriangleup & \simeq \mathbb{E}_{s,z}\left( \left(1-\tanh\left(\hat{y}\left(S,z,\chi,Q,J\right)\right)\right)^{2}\right).\label{eq:scalings_results-1-1-1}
\end{align}
However, different from the case of $\ell_1$-LinR estimator,   there is no analytic solution but
it can be calculated numerically. It can be seen that the $\ell_1$-LinR estimator only differs in the value of scaling factor $\bigtriangleup$ with the $\ell_1$-LogR estimator for Ising model selection.

\section{\label{sec:Replica-symmetric-decoupling}Details of the non-asymptotic result for moderate $M,N$}

As demonstrated in Appendix \ref{Append-sub-Free energy density result}, from the replica analysis, both $\ell_{1}$-LinR and $\ell_{1}$-LogR estimators are decoupled and their asymptotic behavior can be described by two scalar estimators for the active set and inactive set, respectively. It is desirable to obtain the non-asymptotic result for moderate $M,N$. However, it is found that the behavior of the two scalar estimators by simply inserting the finite values of $M,N$ into the EOS does not always lead to good consistency with the experimental results, especially for the $Recall$ when $M$ is small. This can be explained by the derivation of the free energy density. In calculating the energy term $\mathcal{\mathcal{\xi}}$, the limit $M\rightarrow\infty$ is taken
implicitly when assuming the limit $N\rightarrow\infty$ with $\alpha \equiv M/N$. As a result, the scalar estimator associated with the active set can only describe the
asymptotic performance in the limit $M\rightarrow\infty$. Thus, one cannot describe the fluctuating behavior of the estimator in the active set such as the
recall rate for finite $M$. To characterize the non-asymptotic behavior of the estimates in the active set $\Psi$, we replace
the expectation $\mathbb{E}_{s}(\cdot)$  in (\ref{eq:free-energy-result-general}) by the
sample average over $M$ samples, and the corresponding estimates are obtained as 
\begin{equation}
\left\{ \hat{J}_{j}\right\} _{j\in\Psi}=\underset{J_{j,j\in\Psi}}{\arg\min}\left\{ \frac{1}{M}\sum_{\mu=1}^{M}\underset{y^{\mu}}{\min}\left[\frac{\left(y^{\mu}-s_{0}^{\mu}\left(\sqrt{Q}z^{\mu}+\sum_{j\in\Psi}\text{\ensuremath{J_{j}}}s_{j}^{\mu}\right)\right)^{2}}{2\chi}+\ell\left(y^{\mu}\right)\right]+\lambda\sum_{j\in\Psi}\left|J_{j}\right|\right\} ,\label{eq:J-mean-decoupled-result-fluctuate}
\end{equation}
where $z^{\mu}\sim\mathcal{N}\left(0,1\right)$ and $s_{0}^{\mu},s_{j,j\in\Psi}^{\mu}\sim P\left(s_{0},\boldsymbol{s}_{\Psi}|\boldsymbol{J}^{*}\right)$
are random samples $\mu=1,...,M$. Note that the mean estimates $\left\{ \bar{J}_{j}\right\} _{j\in\Psi}$
are replaced by $\left\{ \hat{J}_{j}\right\} _{j\in\Psi}$ in (\ref{eq:J-mean-decoupled-result-fluctuate}) as we now focus on its fluctuating behavior due to the
finite size effect. In the limit $M\rightarrow\infty$, the sample average
will converge to the expectation and thus (\ref{eq:J-mean-decoupled-result-fluctuate})
is equivalent to (\ref{eq:J-mean-decoupled-result}) when $M\rightarrow\infty$. 

\subsection{\label{Appendix-Non-asympt-LASSO}quadratic loss $\ell\left(y\right)=\left(y-1\right)^{2}/2$}

In the case of quadratic loss $\ell\left(y\right)=\left(y-1\right)^{2}/2$,
there is an analytic solution to $y$ in $\underset{y}{\min}\left[\frac{\left(y-s_{0}\left(\sqrt{Q}z+\sum_{j\in\Psi}\text{\ensuremath{\bar{J}_{j}}}s_{j}\right)\right)^{2}}{2\chi}+\ell\left(y\right)\right]$.
Consequently, similar to (\ref{eq:J-mean-result-EOS}), the result of (\ref{eq:J-mean-decoupled-result-fluctuate}) for the $\ell_{1}$-LinR estimator becomes 
\begin{equation}
\left\{ \hat{J}_{j}\right\} _{j\in\Psi} =
\underset{J_{j,j\in\Psi}}{\arg\min}\left[\frac{1}{2\left(1+\chi\right)M}\sum_{\mu=1}^{M}\left(s_{i}^{\mu}-\sum_{j\in\Psi}s_{j}^{\mu}J_{j}-\sqrt{Q}z^{\mu}\right)^{2}+\lambda\sum_{j\in\Psi}\left|J_{j}\right|\right]. 
\label{eq:J-estimate-reduced-1}
\end{equation}
As the mean estimates $\left\{\bar{J}_{j}\right\} _{j\in\Psi}$ are modified as in (\ref{eq:J-estimate-reduced-1}), the corresponding solution to the EOS in (\ref{eq:fixed-point-equations-linear-meanJ}) also needs to be modified, and this can be solved iteratively as sketched in Algorithm \ref{alg:EOS-finite-L1-alog}. For a practical implementation of Algorithm \ref{alg:EOS-finite-L1-alog}, the details are described in the following. 

First, in the EOS (\ref{eq:EOS-linear-meanJ-maintext}), we need to obtain $\tilde{\varLambda}$ satisfying the following relation
\begin{equation}
E\eta=-\int\frac{\rho\left(\gamma\right)}{\tilde{\varLambda}-\gamma}d\gamma,\label{Lambda-equation}
\end{equation}
which is difficult to solve directly. To obtain $\tilde{\varLambda}$, we introduce an auxiliary variable $\varGamma\equiv-\frac{1}{\tilde{\varLambda}}$, by which (\ref{Lambda-equation}) can be rewritten as 
\begin{equation}
\varGamma=\frac{E\eta}{\int\frac{\rho\left(\gamma\right)}{1+\varGamma\gamma}d\gamma},\label{Lambda-equation-equiv}
\end{equation}
which can be solved iteratively. Accordingly, the $\chi,Q,K,H$ in EOS (\ref{eq:EOS-linear-meanJ-maintext}) can be equivalently written in terms of $\varGamma$.

Second, when solving the EOS (\ref{eq:EOS-linear-meanJ-maintext}) iteratively using numerical methods, it is helpful to improve the convergence of the solution by introducing a small amount of damping factor $\mathtt{damp} \in [0,1)$ for $\chi,Q,E,R,F,\eta,K,H,\varGamma$ in each iteration.


The detailed implementation of Algorithm \ref{alg:EOS-finite-L1-alog} is shown in Algorithm \ref{alg:EOS-finite-L1-alog-details}.

\begin{algorithm}[t]
\label{alg:EOS-finite-L1-alog-details}
\DontPrintSemicolon
  \KwInput{$M,N,\lambda,K_0,\rho\left(\gamma\right)$, $\mathtt{damp}, T_{\rm MC}$}
  \KwOutput{$\chi,Q,E,R,F,\eta,K,H,\varGamma,\{\hat{J}^t_{j,j\in\Psi}\}_{t=1}^{T_{MC}}$}  
  \KwInitialize{$\chi,Q,E,R,F,\eta,K,H,\varGamma$}
  \textbf{MC sampling}: {For $t=1...T_{MC}$, draw random samples $s_{0}^{\mu,t},\left\{s^{\mu,t}_{j}\right\}_{j\in\Psi}\sim P\left(s_{0},\boldsymbol{s}_{\Psi}|\boldsymbol{J}^{*}\right)$
and $z^{\mu,t}\sim\mathcal{N}\left(0,1\right)$, $\mu=1...M$}

  \Repeat{convergence}{
    \For{$t=1$ {\bfseries to} $T_{\rm MC}$}{
   Solve $\hat{J}^t_{j,j\in\Psi} = \underset{J_{j,j\in\Psi}}{\arg\min}\left[\frac{\sum_{\mu=1}^{M}\left(s_{0}^{\mu,t}-\sum_{j\in\Psi}s_{j}^{\mu,t}J_{j}-\sqrt{Q}z^{\mu,t}\right)^{2}}{2\left(1+\chi\right)M}+\lambda\sum_{j\in\Psi}\left|{J}_{j}\right|\right]$
    
   Compute $\triangle\left(t\right)=\frac{1}{M}\sum_{\mu=1}^{M}\left(s_{0}^{\mu,t}-\sum_{j\in\Psi}s_{j}^{\mu,t}\hat{J}_{j}^t\right)^{2}$
    }
   Set $\bar{\triangle} =\frac{1}{T_{\rm MC}}\sum_{t_1=1}^{T_{\rm MC}}\triangle\left(t\right)$
   
   $E=(1 - \mathtt{damp})\frac{\alpha}{\left(1+\chi\right)} + \mathtt{damp}\cdot{E}$ 
   
    $F=(1 - \mathtt{damp})\frac{\alpha}{\left(1+\chi\right)^{2}}\left(\bar{\triangle}+Q\right)+\mathtt{damp}\cdot{F}$
    
    $R = (1 - \mathtt{damp})\frac{1}{K^{2}}\left[\left(H+\frac{\lambda^{2}M^{2}}{N}\right)\textrm{erfc}\left(\frac{\lambda M}{\sqrt{2HN}}\right)-2\lambda M\sqrt{\frac{H}{N}}\frac{1}{\sqrt{2\pi}}e^{-\frac{\lambda^{2}M^{2}}{2HN}}\right] + \mathtt{damp}\cdot{R}$
    
    \Repeat{convergence}{
    $\varGamma = (1 - \mathtt{damp})\frac{E\eta}{\int\frac{\rho\left(\gamma\right)}{1+\varGamma\gamma}d\gamma} + \mathtt{damp}\cdot{\varGamma}$}
    
    $K=(1 - \mathtt{damp})\left(-\frac{E}{\varGamma}+\frac{1}{\eta}\right) + \mathtt{damp}\cdot{K}$
    
    $\chi=(1 - \mathtt{damp})\left(-\frac{\eta}{\varGamma}+\frac{1}{E}\right) + \mathtt{damp}\cdot{\chi}$  
    
    $Q=(1 - \mathtt{damp})\left(\frac{F}{E^{2}}-\frac{R}{\varGamma} -\frac{\left(-ER+F\eta\right)\eta}{\varGamma^2\int\frac{\rho\left(\gamma\right)}{\left(1+\varGamma\gamma\right)^{2}}d\gamma}\right) + \mathtt{damp}\cdot{Q}$
    
   $H=(1 - \mathtt{damp})\left(\frac{R}{E^{2}}-\frac{F}{\varGamma} -\frac{\left(-ER+F\eta\right)E}{\varGamma^2\int\frac{\rho\left(\gamma\right)}{\left(1+\varGamma\gamma\right)^{2}}d\gamma}\right) + \mathtt{damp}\cdot{H}$
   
   $\eta=(1 - \mathtt{damp})\frac{1}{K}\textrm{erfc}\left(\frac{\lambda M}{\sqrt{2HN}}\right) +  \mathtt{damp}\cdot{\eta}$
   }
\caption{Detailed implementation of Algorithm \ref{alg:EOS-finite-L1-alog} for the $\ell_{1}$-LinR estimator with moderate $M,N$.}
\end{algorithm}

\subsection{Logistic loss $\ell\left(y\right)=\log\left(1+e^{-2y}\right)$}

In the case of square lass $\ell\left(y\right)=\log\left(1+e^{-2y}\right)$,
since there is no analytic solution to $y$ in $\underset{y}{\min}\left[\frac{\left(y-s_{0}\left(\sqrt{Q}z+\sum_{j\in\Psi}\text{\ensuremath{\bar{J}_{j}}}s_{j}\right)\right)^{2}}{2\chi}+\ell\left(y\right)\right]$, the result of (\ref{eq:J-mean-decoupled-result-fluctuate}) for the $\ell_{1}$-LogR estimator becomes 
\begin{equation}
\hat{J}_{j,j\in\Psi}=
\underset{J_{j,j\in\Psi}}{\arg\min}\left[\frac{1}{M}\sum_{\mu=1}^{M}\underset{y^{\mu}}{\min}\left[\frac{\left(y^{\mu}-s_{0}^{\mu}\left(\sqrt{Q}z^{\mu}+\sum_{j\in\Psi}\text{\ensuremath{J_{j}}}s_{j}^{\mu}\right)\right)^{2}}{2\chi}+\log\left(1+e^{-2y}\right)\right]+\lambda\sum_{j\in\Psi}\left|J^{\mu}_{j}\right|\right],
\label{eq:J-estimate-reduced-LR}
\end{equation}
Similarly as the case for quadratic loss, as the mean estimates $\left\{\bar{J}_{j}\right\} _{j\in\Psi}$ are modified as in (\ref{eq:J-estimate-reduced-LR}), the corresponding solutions to the EOS in (\ref{eq:fixed-point-equations-logistic-meanJ}) also need to be modified, which can be solved iteratively as shown in Algorithm \ref{alg:EOS-finite-LR-details}. 

\begin{algorithm}[t]
\label{alg:EOS-finite-LR-details}
\DontPrintSemicolon
  \KwInput{$M,N,\lambda,K_0,\rho\left(\gamma\right)$, $\mathtt{damp}, T_{\rm MC}$}
  \KwOutput{$\chi,Q,E,R,F,\eta,K,H,\varGamma,\{\hat{J}^t_{j,j\in\Psi}\}_{t=1}^{T_{MC}}$}  
  \KwInitialize{$\chi,Q,E,R,F,\eta,K,H,\varGamma$}
  
  \textbf{MC sampling}: {For $t=1...T_{MC}$, draw random samples $s_{0}^{\mu,t},\left\{s^{\mu,t}_{j}\right\}_{j\in\Psi}\sim P\left(s_{0},\boldsymbol{s}_{\Psi}|\boldsymbol{J}^{*}\right)$
and $z^{\mu,t}\sim\mathcal{N}\left(0,1\right)$, $\mu=1...M$}

  \Repeat{convergence}{
   \For{$t=1$ {\bfseries to} $T_{\rm MC}$}{
    Initialization $\hat{J}^t_{j,j\in\Psi}$
    
   \Repeat{convergence}{
   \small{$\hat{y}^{\mu,t} = \underset{y^{\mu,t}}{\arg\min}\left[\frac{\left(y^{\mu}-s_{0}^{\mu,t}\left(\sqrt{Q}z^{\mu,t}+\sum_{j\in\Psi}\text{\ensuremath{\hat{J}_{j}}}s_{j}^{\mu,t}\right)\right)^{2}}{2\chi}+\log\left(1+e^{-2y^{\mu}}\right)\right],\mu=1...M$}
   
    \small{$\hat{J}^t_{j,j\in\Psi}=\underset{J_{j,j\in\Psi}}{\arg\min}\bigg\{\frac{1}{M}\sum_{\mu=1}^{M}\left[\frac{\left(\hat{y}^{\mu,t}-s_{0}^{\mu,t}\left(\sqrt{Q}z^{\mu,t}+\sum_{j\in\Psi}\text{\ensuremath{J^t_{j}}}s_{j}^{\mu,t}\right)\right)^{2}}{2\chi}+\log\left(1+e^{-2\hat{y}^{\mu,t}}\right)\right]+\hspace{3cm}\lambda\sum_{j\in\Psi}\left|J_{j}\right|\bigg\}$
    }}

    Compute $\triangle_1\left(t\right) =\frac{1}{M}\sum_{\mu=1}^{M}\left(\frac{s^{\mu,t}_{0}z^{\mu,t}}{-\sqrt{Q}}(1-\tanh\left(\hat{y}^{\mu,t}\right))\right)$
    
   Compute $\triangle_2\left(t\right) =\frac{1}{M}\sum_{\mu=1}^{M}\left(1-\tanh\left(\hat{y}^{\mu,t}\right)\right)^{2}$}

   Set $\bar{\triangle}_1 =\frac{1}{T_{\rm MC}}\sum_{t=1}^{T_{\rm MC}}\triangle_1\left(t\right)$ and $\bar{\triangle}_2 =\frac{1}{T_{\rm MC}}\sum_{t=1}^{T_{\rm MC}}\triangle_2\left(t\right)$

    $E=(1 - \mathtt{damp})\cdot{\alpha\bar{\triangle}_1} + \mathtt{damp}\cdot{E}$ 
    
   $F=(1 - \mathtt{damp})\cdot{\alpha\bar{\triangle}_2} +\mathtt{damp}\cdot{F}$
   
   $R = (1 - \mathtt{damp})\frac{1}{K^{2}}\left[\left(H+\frac{\lambda^{2}M^{2}}{N}\right)\textrm{erfc}\left(\frac{\lambda M}{\sqrt{2HN}}\right)-2\lambda M\sqrt{\frac{H}{N}}\frac{1}{\sqrt{2\pi}}e^{-\frac{\lambda^{2}M^{2}}{2HN}}\right] + \mathtt{damp}\cdot{R}$
   
   \Repeat{convergence}{
    $\varGamma = (1 - \mathtt{damp})\frac{E\eta}{\int\frac{\rho\left(\gamma\right)}{1+\varGamma\gamma}d\gamma} + \mathtt{damp}\cdot{\varGamma}$}

    $K=(1 - \mathtt{damp})\left(-\frac{E}{\varGamma}+\frac{1}{\eta}\right) + \mathtt{damp}\cdot{K}$
   
    $\chi=(1 - \mathtt{damp})\left(-\frac{\eta}{\varGamma}+\frac{1}{E}\right) + \mathtt{damp}\cdot{\chi}$  
  
   $Q=(1 - \mathtt{damp})\left(\frac{F}{E^{2}}-\frac{R}{\varGamma} -\frac{\left(-ER+F\eta\right)\eta}{\varGamma^2\int\frac{\rho\left(\gamma\right)}{\left(1+\varGamma\gamma\right)^{2}}d\gamma}\right) + \mathtt{damp}\cdot{Q}$
  
   $H=(1 - \mathtt{damp})\left(\frac{R}{E^{2}}-\frac{F}{\varGamma} -\frac{\left(-ER+F\eta\right)E}{\varGamma^2\int\frac{\rho\left(\gamma\right)}{\left(1+\varGamma\gamma\right)^{2}}d\gamma}\right) + \mathtt{damp}\cdot{H}$
 
  $\eta=(1 - \mathtt{damp})\frac{1}{K}\textrm{erfc}\left(\frac{\lambda M}{\sqrt{2HN}}\right) +  \mathtt{damp}\cdot{\eta}$}

\caption{Detailed implementation of
   solving the EOS (\ref{eq:fixed-point-equations-logistic-meanJ}) together with (\ref{eq:J-estimate-reduced-LR}) for  $\ell_{1}$-LogR  with moderate $M,N$.}
\end{algorithm}

\section{\label{subsec:Eigenvalue-distribution}Eigenvalue Distribution $\rho\left(\gamma\right)$}
From the replica analysis presented, the learning performance will depend
on the eigenvalue distribution (EVD) $\rho\left(\gamma\right)$ of the
covariance matrix $C$ of the original Ising model.

\textcolor{black}{There are two issues to be noted. One is about the formula connecting the performance of the estimator and the spectral density, and the other is the numeric values of quantities which are computed from the formula. For the first point, no assumption about the spectral density is needed to obtain the formula itself and this formula is valid when the graph structure is tree-like and the Ising model defined on the graph is in the paramagnetic phase. For the second point, we need the specific form of the spectral density to obtain numeric solutions in general. As a demonstration, we assume the random regular graph with constant coupling strength for which the spectral density can be obtained analytically as has already been known before in \cite{Abbara2019c}. }
 
In general, it is difficult to obtain this EVD; however, for sparse tree-like graphs such as RR graph $G\in \mathcal{G}_{N,d,K_{0}}$ with constant node degree
$d$ and sufficiently small coupling strength $K_{0}$ that yields the paramagnetic state ($\mathbb{E}_{\boldsymbol{s}}(\boldsymbol{s}) =\boldsymbol{0}$), it can be computed analytically. For this, we express the covariances as 
\begin{equation}
C_{ij} = \mathbb{E}_{\boldsymbol{s}}(s_i s_j) -\mathbb{E}_{\boldsymbol{s}}(s_i) \mathbb{E}_{\boldsymbol{s}}(s_j)
= \frac{\partial^2 \log Z(\boldsymbol{\theta}) }
{\partial \theta_i \partial \theta_j}, 
\end{equation}
where 
$Z(\boldsymbol{\theta}) = \int d\boldsymbol{s} P_{\rm Ising}(\boldsymbol{s}|J^*) 
\exp(\sum_{i=0}^{N-1} \theta_i s_i) $ and the assessment is carried out at $\boldsymbol{\theta} = \boldsymbol{0}$. 

In addition, for technical convenience we introduce the Gibbs free energy  as
\begin{equation}
A\left(\boldsymbol{m}\right)=\underset{\boldsymbol{\theta}}{\max}\left\{ \boldsymbol{\theta}^{T}\boldsymbol{m}-\log Z\left(\boldsymbol{\theta}\right)\right\}.\label{eq:Gibbs-Gm-1}
\end{equation}
The definition of (\ref{eq:Gibbs-Gm-1}) indicates that following two  relations hold: 
\begin{align}
\frac{\partial m_i}{\partial \theta_j} 
=\frac{\partial^2 \log Z(\boldsymbol{\theta})}{\partial \theta_i \partial \theta_j} = C_{ij}, \nonumber \\
\frac{\partial \theta_i}{\partial m_j} =[C^{-1}]_{ij} 
= \frac{\partial^2 A(\boldsymbol{m})}{\partial m_i
\partial m_j},
\end{align}
where the evaluations are performed at $\boldsymbol{\theta} = \boldsymbol{0}$ and $\boldsymbol{m} = \arg\min _{\boldsymbol{m}} A(\boldsymbol{m})$ ($=\boldsymbol{0}$ under the paramagnetic assumption).

Consequently, we can focus on the computation of $A\left(\boldsymbol{m}\right)$
to obtain the EVD of $C^{-1}$. The inverse covariance matrix of a RR graph $G\in \mathcal{G}_{N,d,K_{0}}$ can be computed
from the Hessian of the Gibbs free energy \cite{Abbara2019c, ricci2012bethe, nguyen2012bethe} as 
\begin{align}
\left[C^{-1}\right]_{ij} & =\frac{\partial A\left(\boldsymbol{m}\right)}{\partial m_{i}\partial m_{j}}\nonumber \\
 & =\left(\frac{d}{1-\tanh^{2}K_0}-d+1\right)\delta_{ij}-\frac{\tanh\left(J_{ij}\right)}{1-\tanh^{2}\left(J_{ij}\right)}\left(1-\delta_{ij}\right),\label{eq:ivnerse-correlation-mat-result}
\end{align}
and in matrix form, we have
\begin{equation}
C^{-1}=\left(\frac{d}{1-\tanh^{2}K_0}-d+1\right)\mathbf{I}-\frac{\tanh\left(\boldsymbol{J}\right)}{1-\tanh^{2}\left(\boldsymbol{J}\right)}, \label{eq:inverse-corre-mat-C-1}
\end{equation}
where $\mathbf{I}$ is an identity matrix of proper size, and the operations $\tanh\left(\cdot\right), \tanh^2\left(\cdot\right)$ on matrix $\boldsymbol{J}$ are defined in the component-wise manner. For RR graph $G\in \mathcal{G}_{N,d,K_{0}}$, $\boldsymbol{J}$ is a sparse matrix, therefore the matrix $\frac{\tanh\left(\boldsymbol{J}\right)}{1-\tanh^{2}\left(\boldsymbol{J}\right)}$ also corresponds to a sparse coupling matrix (whose nonzero coupling positions are the same as $\boldsymbol{J}$) with constant coupling strength $K_{1}=\frac{\tanh\left(K_0\right)}{1-\tanh^{2}\left(K_0\right)}$ and fixed connectivity $d$, the corresponding
eigenvalue (denoted as $\zeta$) distribution can be calculated as \citep{mckay1981expected}
\begin{equation}
\rho_{\zeta}\left(\zeta\right)=\frac{d\sqrt{4K_{1}^{2}\left(d-1\right)-\zeta^{2}}}{2\pi\left(K_{1}^{2}d^{2}-\zeta^{2}\right)},\;\left|\zeta\right|\leq2K_{1}\sqrt{d-1}. \label{eq:EVD-kesi}
\end{equation}
From (\ref{eq:inverse-corre-mat-C-1}), the eigenvalue $\eta$
of $C^{-1}$ is
\begin{equation}
\eta_{i}=\frac{d}{1-\tanh^{2}K_0}-d+1-\zeta_{i},\label{eq:eigen-value-result}
\end{equation}
which, when combined with (\ref{eq:EVD-kesi}), 
readily yields the EVD of $\eta$ as $N\rightarrow\infty$ as follows:
\begin{align}
\rho_{\eta}\left(\eta\right) & =\rho_{\zeta}\left(\frac{d}{1-\tanh^{2}K_0}-d+1-\eta\right)\nonumber \\
 & =\frac{d\sqrt{4\left(\frac{\tanh\left(K_0\right)}{1-\tanh^{2}\left(K_0\right)}\right)^{2}\left(d-1\right)-\left(\frac{d}{1-\tanh^{2}K_0}-d+1-\eta\right)^{2}}}{2\pi\left(\left(\frac{\tanh\left(K_0\right)}{1-\tanh^{2}\left(K_0\right)}\right)^{2}d^{2}-\left(\frac{d}{1-\tanh^{2}K_0}-d+1-\eta\right)^{2}\right)},\label{eq:EVD-nita-results}
\end{align}
where $\eta\in\left[\frac{d}{1-\tanh^{2}K_0}-d+1-\frac{2\tanh\left(K_0\right)\sqrt{d-1}}{1-\tanh^{2}\left(K_0\right)},\frac{d}{1-\tanh^{2}K_0}-d+1+\frac{2\tanh\left(K_0\right)\sqrt{d-1}}{1-\tanh^{2}\left(K_0\right)}\right]$.

Consequently, since  $\gamma=1/\eta$, we obtain the EVD of $\rho\left(\gamma\right)$ as follows
\begin{align}
\rho\left(\gamma\right) & =\frac{1}{\gamma^2} \rho_{\eta}\left(\eta=\frac{1}{\gamma}\right) \nonumber \\
& =\frac{d\sqrt{4\left(\frac{\tanh\left(K_0\right)}{1-\tanh^{2}\left(K_0\right)}\right)^{2}\left(d-1\right)-\left(\frac{d}{1-\tanh^{2}K_0}-d+1-\frac{1}{\gamma}\right)^{2}}}{2\pi\gamma^2\left(\left(\frac{\tanh\left(K_0\right)}{1-\tanh^{2}\left(K_0\right)}\right)^{2}d^{2}-\left(\frac{d}{1-\tanh^{2}K_0}-d+1-\frac{1}{\gamma}\right)^{2}\right)}
\end{align}
where $\gamma\in\left[1/\left(\frac{d}{1-\tanh^{2}K_0}-d+1+\frac{2\tanh\left(K_0\right)\sqrt{d-1}}{1-\tanh^{2}\left(K_0\right)}\right),1/\left(\frac{d}{1-\tanh^{2}K_0}-d+1-\frac{2\tanh\left(K_0\right)\sqrt{d-1}}{1-\tanh^{2}\left(K_0\right)}\right)\right]$.

\section{\label{appedix:additional results}Additional Experimental Results}
\begin{figure*}[htbp]
\advance\leftskip -1cm
\includegraphics[width=16cm]{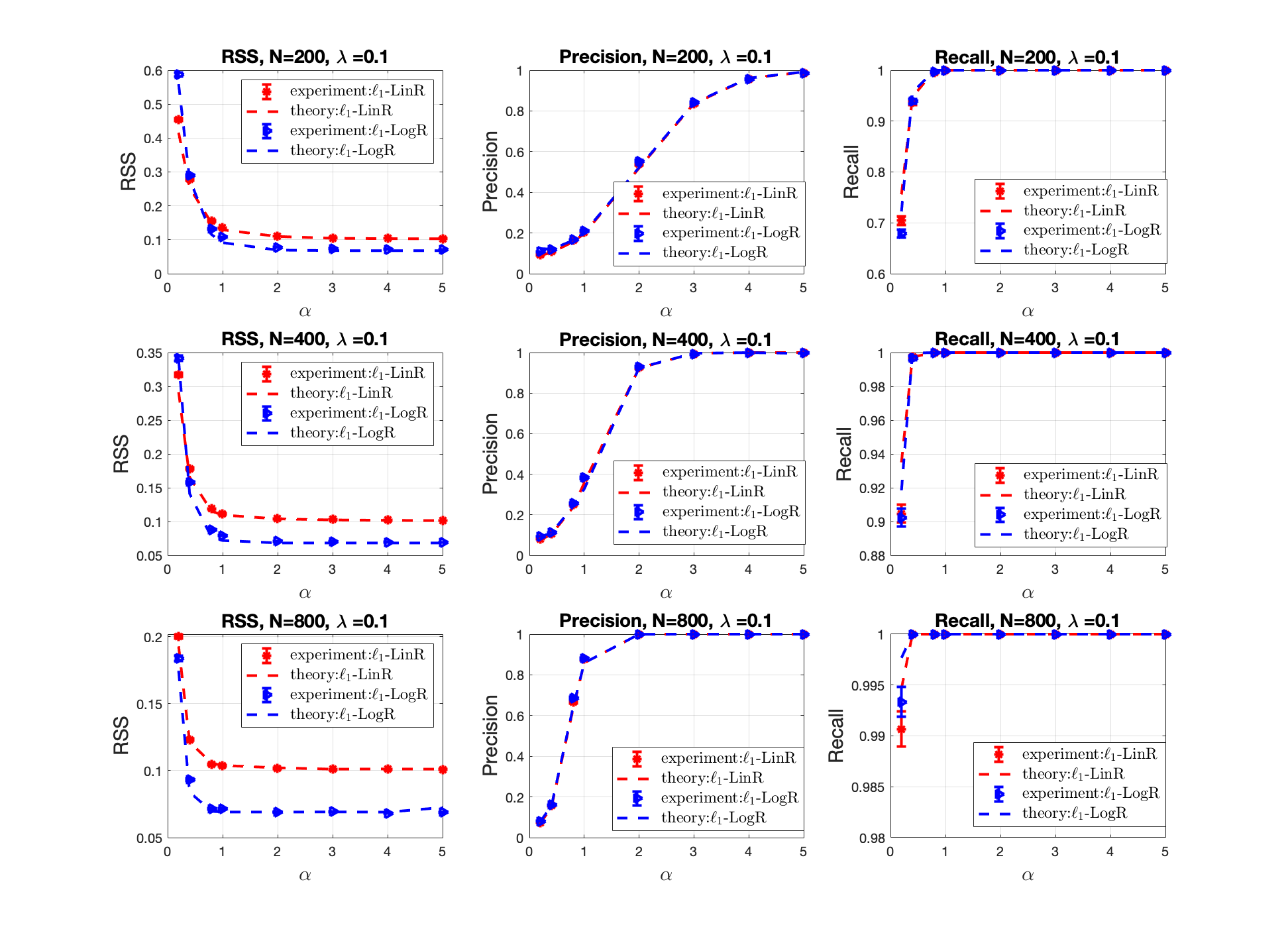}\caption{Theoretical and experimental results of $RSS$, $Precision$ and $Recall$ for both $\ell_{1}$-LinR and $\ell_{1}$-LogR
when $\lambda=0.1$, $N=200,400,800$ with different values of $\alpha \equiv M/N$. The standard error bars are obtained from 1000 random runs. 
An excellent agreement between theory  and
experiment is achieved, even for small $N=200$ and small $\alpha$
( small $M$).\label{fig:RSS-Precision-Recall-N200N400N800-lambda01}}
\end{figure*}

\begin{figure*}
\advance\leftskip -1cm
\includegraphics[width=16cm]{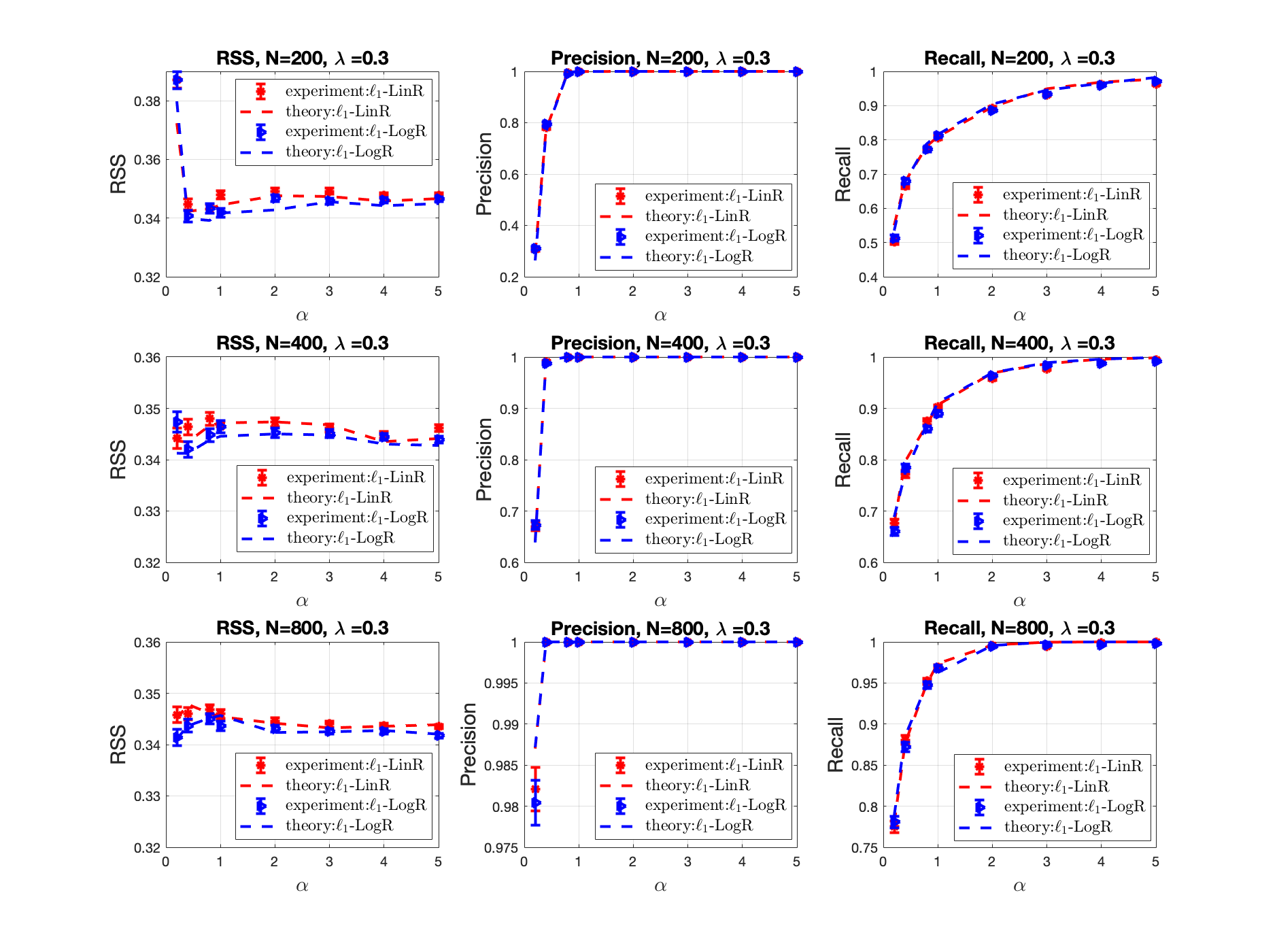}\caption{Theoretical and experimental results of RSS, $Precision$ and $Recall$ for both $\ell_{1}$-LinR and $\ell_{1}$-LogR
when $\lambda=0.3$, $N=200,400,800$ with different values of $\alpha \equiv M/N$. The standard error bars are obtained from 1000 random runs. 
An excellent agreement between theory  and
experiment is achieved, even for small $N=200$ and small $\alpha$
( small $M$).\label{fig:J-Q-R-N200N400}}
\end{figure*}

Fig. \ref{fig:RSS-Precision-Recall-N200N400N800-lambda01}
and Fig. \ref{fig:J-Q-R-N200N400}   show the full results of non-asymptotic learning performance prediction when  $\lambda=0.1$ and $\lambda=0.3$, respectively. Good agreements between replica results and experimental
results are achieved in all cases. As can be seen, there is negligible difference
in $Precision$ and $Recall$ between $\ell_{1}$-LinR and $\ell_{1}$-LogR.
Meanwhile, compared to Fig. \ref{fig:RSS-Precision-Recall-N200N400N800-lambda01}
when $\lambda=0.1$, the difference in RSS between $\ell_{1}$-LinR
and $\ell_{1}$-LogR is reduced when $\lambda=0.3$. In addition,
by comparing Fig. \ref{fig:RSS-Precision-Recall-N200N400N800-lambda01}
and Fig. \ref{fig:J-Q-R-N200N400}, it can be seen that under the
same setting, when $\lambda$ increases, the $Precision$ becomes
larger while the $Recall$ becomes smaller, implying a tradeoff in choosing $\lambda$ in practice for Ising model selection with finite $M,N$.

\begin{figure}[htbp]
\centering 
\includegraphics[width=12cm]{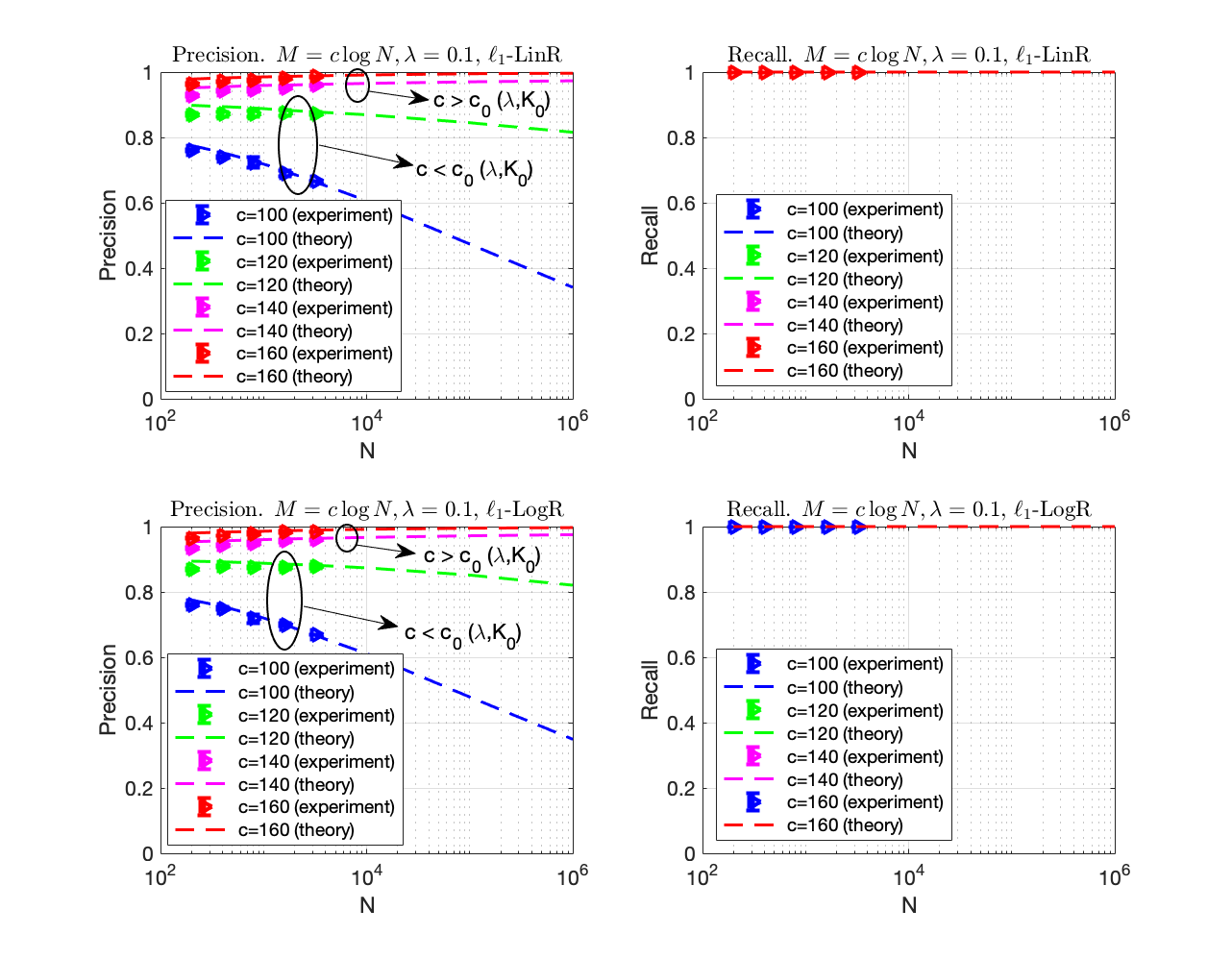}\caption{$Precision$ and $Recall$ versus $N$ when $M=c\log N$ and $K_{0}=0.4$
for $\ell_{1}$-LinR and $\ell_{1}$-LogR when $\lambda=0.1$, where $c_{0}\left(\lambda,K_{0}\right)\equiv\frac{c\left(\lambda,K_{0}\right)}{\lambda^{2}}\approx137$. When $c>c_{0}\left(\lambda,K_{0}\right)$,
the \textit{Precision }increases consistently with $N$ and approaches
1 as $N\rightarrow\infty$ while it decreases consistently with $N$
when $c<c_{0}\left(\lambda,K_{0}\right)$. \label{fig:Precision-recall-aroundC}}
\end{figure}

\begin{figure}[htbp]
\centering \includegraphics[width=12cm]{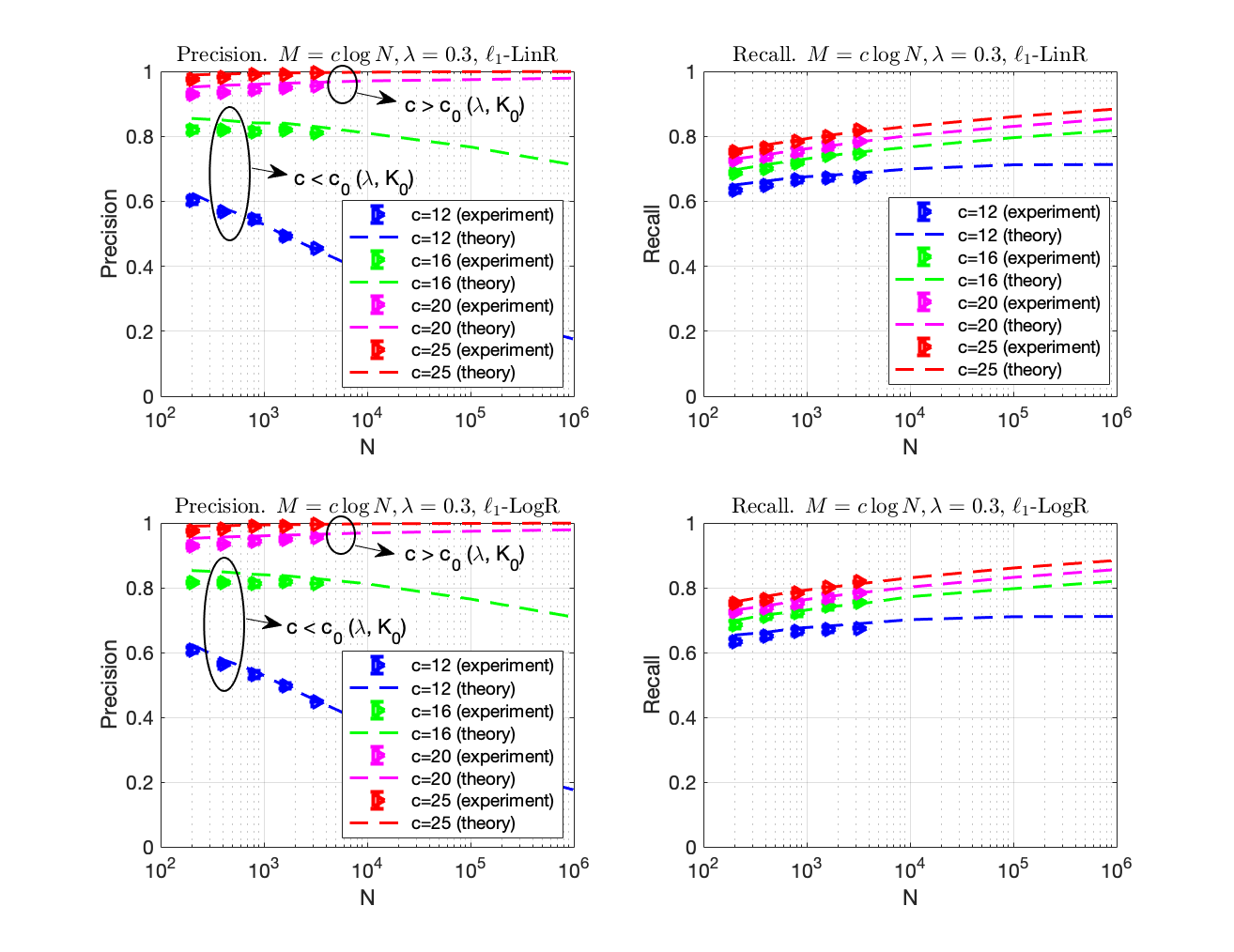}
\vspace{-1em}
\caption{$Precision$ and $Recall$ versus $N$ when $M=c\log N$ and $K_{0}=0.4$
for $\ell_{1}$-LinR and $\ell_{1}$-LogR when $\lambda=0.3$, where $c_{0}\left(\lambda,K_{0}\right)\equiv\frac{c\left(\lambda,K_{0}\right)}{\lambda^{2}}\approx19.4$. When $c>c_{0}\left(\lambda,K_{0}\right)$,
the \textit{Precision }increases consistently with $N$ and approaches
1 as $N\rightarrow\infty$ while it decreases consistently with $N$
when $c<c_{0}\left(\lambda,K_{0}\right)$. The \textit{Recall} increases
consistently and approach to 1 as $N\rightarrow\infty$. \label{fig:Precision-recall-aroundC-1}}
\end{figure}

Fig. \ref{fig:Precision-recall-aroundC} and Fig. \ref{fig:Precision-recall-aroundC-1} show the full results of critical scaling  prediction when  $\lambda=0.1$ and $\lambda=0.3$, respectively. For comparison, both the results of  $\ell_{1}$-LinR and $\ell_{1}$-LogR are shown. It can be seen that apart from the good agreements between replica results and experimental
results, the prediction of the scaling value $c_{0}\left(\lambda,K_{0}\right)\equiv\frac{c\left(\lambda,K_{0}\right)}{\lambda^{2}}$ is very accurate.

\end{document}